\PassOptionsToPackage{dvipsnames,table}{xcolor}
\documentclass[10pt,twocolumn,letterpaper]{article}

\usepackage[final]{iccv}
\definecolor{iccvblue}{rgb}{0.21,0.49,0.74}
\usepackage[pagebackref,breaklinks,colorlinks,allcolors=iccvblue]{hyperref}

\usepackage[accsupp]{axessibility}
\usepackage{xcolor}
\usepackage{multirow}
\usepackage{array}
\usepackage{tabularray}
\usepackage{tabularx}
\usepackage{multicol}
\usepackage{ifthen}
\usepackage{float}
\usepackage{subfiles}

\def\paperID{7343}
\def\confName{ICCV}
\def\confYear{2025}

\def\makeappendix
{
\onecolumn
\appendix
\section*{Appendix}
\addcontentsline{toc}{section}{Appendix}
\setcounter{section}{0}
\renewcommand{\thesection}{\Alph{section}}
}

\newcommand{\figref}[1]{Fig.~\ref{#1}}
\newcommand{\Figref}[1]{Figure~\ref{#1}}

\newcommand{\Tabref}[1]{Table~\ref{#1}}

\newcommand{\secref}[1]{Sec.~\ref{#1}}

\newcommand{\OurDataset}{ROVI}

\makeatletter 
\newcommand{\compare}[3][,]{%
  \def\leftformat{}%
  \def\rightformat{}%
  \def\itemcount{0}%
  \@for\@option:=#1\do{%
    \ifnum\itemcount=0%
      \edef\leftformat{\@option}%
      \def\itemcount{1}%
    \else%
      \edef\rightformat{\@option}%
    \fi%
  }%
  \ifthenelse{\equal{\leftformat}{bold}}{\textbf{#2}}{#2}%
  \textcolor{brown}
  {/%
    \ifthenelse{\equal{\rightformat}{bold}}{\textbf{#3}}{#3}%
  }%
}
\makeatother

\title{\OurDataset: A VLM-LLM Re-Captioned Dataset for Open-Vocabulary Instance-Grounded Text-to-Image Generation}

\author{Cihang Peng \quad\quad Qiming Hou \quad\quad Zhong Ren \quad\quad Kun Zhou\footnotemark[1]\\
State Key Lab of CAD\&CG, Zhejiang University\\
{\tt\small cihangpeng@zju.edu.cn, hqm03ster@gmail.com, renzhong@zju.edu.cn, kunzhou@acm.org}
}

\begin{document}

\twocolumn[{%
\renewcommand\twocolumn[1][]{#1}%
\maketitle
\vspace{-15pt}
\centering
\includegraphics[width=1.0\linewidth]{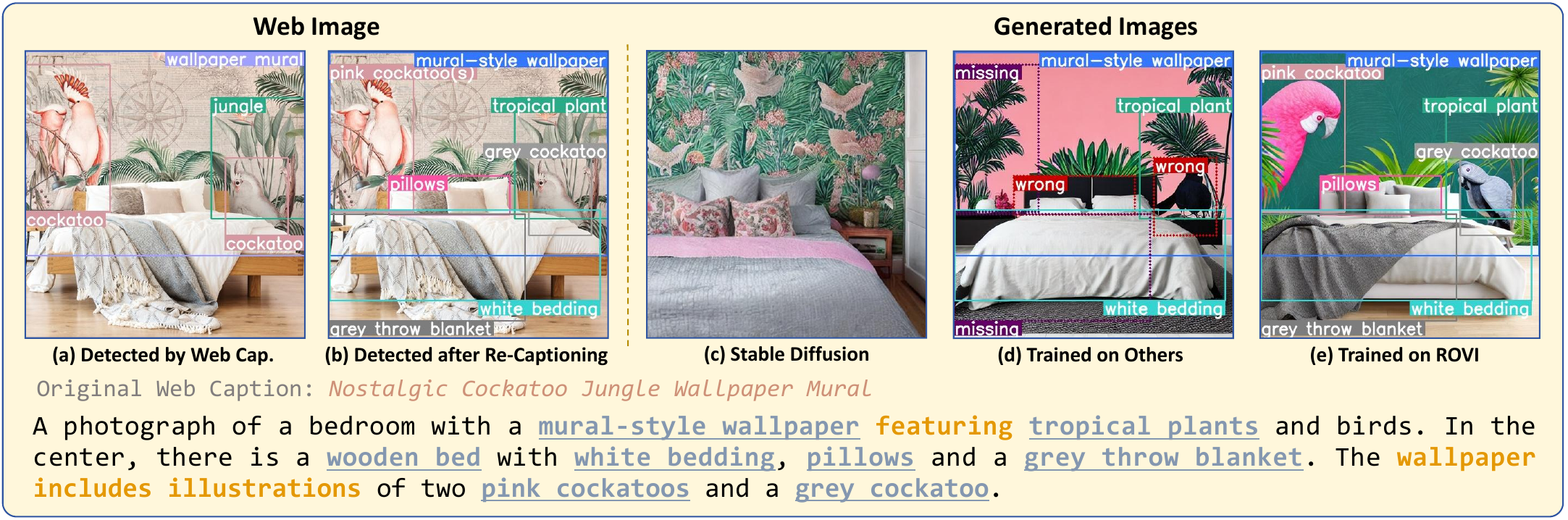}
\vspace{-18pt}
\captionof{figure}{
Effect of our pre-detection strategy. Image (a) shows a web image with original caption where OVDs only detect \emph{cockatoo, jungle, and wallpaper mural}. Our method (b) discovers additional elements: \emph{bedding, pillows, and blanket}, with detailed composition descriptions. The base generator~\cite{stablediffusion} (c) captures composition but lacks instance grounding. The instance-grounded model~\cite{GLIGEN} (d) fine-tuned on sparse web captions only generates a bird-like shadow, omitting the blanket and other elements. The same algorithm trained on our re-captioned dataset (e) correctly produces all details.
\label{fig:introcmp}
}
\vspace{5pt}
}]

\begin{abstract}
We present \OurDataset{}, a high-quality synthetic dataset for instance-grounded text-to-image generation, created by labeling 1M curated web images. Our key innovation is a strategy called re-captioning, focusing on the pre-detection stage, where a VLM (Vision-Language Model) generates comprehensive visual descriptions that are then processed by an LLM (Large Language Model) to extract a flat list of potential categories for OVDs (Open-Vocabulary Detectors) to detect. 
This approach yields a global prompt inherently linked to instance annotations while capturing secondary visual elements humans typically overlook. 
Evaluations show that \OurDataset{} exceeds existing detection datasets in image quality and resolution while containing two orders of magnitude more categories with an open-vocabulary nature. For demonstrative purposes, a text-to-image model GLIGEN~\cite{GLIGEN} trained on \OurDataset{} significantly outperforms state-of-the-art alternatives in instance grounding accuracy, prompt fidelity, and aesthetic quality. 
Our dataset and reproducible pipeline are available at~\url{https://github.com/CihangPeng/ROVI}.
\end{abstract}

\begin{table*}[t]
    \centering
    \renewcommand{\arraystretch}{1.05} 
    
    \begin{tabularx}{\textwidth}{c|cccccc}
        \hline
        \textbf{Dataset} & \textbf{Images} & \textbf{Categories} & \textbf{Avg. Category} & \textbf{Avg. Box} & \textbf{Avg. Resolution} & \textbf{Avg. Aesthetic Score}\\ 
        \hline
        COCO~\cite{COCO} & 123K & 80 & 2.93 & 7.04 & $578\times484$ & 4.97 \\
        LVIS~\cite{LVIS} & 123K & 1203 & 3.62 & 12.73 & $578\times484$ & 4.97 \\
        Objects365~\cite{obj365} & 1936K & 365 & 6.12 & 14.58 & $663\times536$ & 4.92 \\
        Open Images~\cite{openimages} & 1893K & 600 & 2.62 & 8.38 & $976\times791$ & 4.90 \\
        V3det~\cite{v3det} & 245K & 13204 & 4.01 & 7.16 & $789\times660$ & 4.99 \\
        \hline
        \OurDataset{} & 1012K & \textbf{1443360} & \textbf{12.51} & \textbf{24.21} & $\textbf{2102}\times\textbf{1488}$ & \textbf{6.00}  \\
        \hline
    \end{tabularx}
    \vspace{-5pt}
    \caption{Dataset statistics computed on combined training and validation sets. ``Categories" counts the total number of distinct categories in the entire dataset. ``Avg. Category" counts the average number of distinct categories in each image.}
    \label{tab:datasetsCmp}
    \vspace{-10pt}
\end{table*}

\section{Introduction}\label{sec:intro}
\vspace{-3pt}
Text-to-image generation has rapidly matured in recent years with huge industrial impact. While in basic form it solely takes a text prompt as input, many practical applications would benefit from \emph{instance grounding}, 
where users specify precisely where and what kind of object should appear using box-label pairs to enable finer control. 
Many state-of-the-art research~\cite{GLIGEN,MIGC,MIGCpp,InstanceDiffusion,densediffusion,multidiffusion,ifadapter} can provide this capability, giving significant insight on algorithm design. Unfortunately, their performance is restricted by the poor availability of dedicated \emph{datasets}.

\renewcommand{\thefootnote}{\fnsymbol{footnote}}
\footnotetext[1]{Corresponding author}

Typically, the training of instance-grounded generation models consumes images with individually labeled object instances with bounding boxes and captions. 
Existing public datasets like COCO~\cite{COCO} and LVIS~\cite{LVIS} provide annotations that superficially meet these requirements but suffer from critical limitations. 
First, these datasets are primarily designed for detection purposes, thus cannot guarantee high-quality text-image pairs. The image descriptions are either missing or of poor quality with insufficient detail, while many images feature poor lighting conditions, heavy occlusion, and cluttered backgrounds, all of which adversely affect text-to-image generation quality. Second, their label vocabulary remain restricted to a limited set of predefined categories, severely constraining the expressive potential of instance grounding tasks.
Recent algorithm research on instance grounding ~\cite{GLIGEN,MIGC,MIGCpp,InstanceDiffusion,ifadapter,ranni} tend to generate \textit{ad hoc} synthetic instance labels to enrich the training data. Such dataset synthesis is enabled by the recent advancement of OVDs (Open-Vocabulary Detectors)~\cite{owlv2,ovdino,yoloworld,groundingdino,detclipv3,cvprovd1,cvprovd2}. While OVDs can detect arbitrary objects mentioned in the accompanying text, captions rarely include all visual elements, and the detection often remains incomplete. \Figref{fig:introcmp} illustrates how such incompletion eventually translates to a low-quality generated image: a web image of a cockatoo mural in (a) has a caption mentioning only the cockatoo, omitting visible elements such as bed, pillows, and blanket. Using a model trained on the resulting annotations, the generated images fail to properly integrate previously ``unseen" objects, as shown in (d).

To address this limitation, we introduce \OurDataset{} (Re-captioned Open-Vocabulary Instances), a high-quality synthetic dataset of 1M images created through novel methodology. Our core innovation is a pre-detection re-captioning strategy with two main components. First, we use a VLM (Vision-Language Model) to create a detailed description of recognizable objects for each image along with compositional information such as semantic relationships between instances. Then we use an LLM (Large Language Model) to summarize a flat list of categories from the VLM description. The end result is illustrated in \figref{fig:introcmp}~(b) and the accompanying text. Various components of the bed are detected and the birds are explicitly stated as a part of the wallpaper in the caption. Such improved dataset eventually leads to a network that can correctly interpret the instance layout in \figref{fig:introcmp}~(e).

\Tabref{tab:datasetsCmp} compares the overall statistics with existing alternatives. Our \OurDataset{} dataset is able to discover two orders of magnitude more distinct object categories, producing significantly more grounded instances for each image. We also curated the dataset with a considerably higher standard in terms of resolution and aesthetic scores.

Our dataset and codes are publicly available, including image URLs, detailed annotations, and intermediate results from VLM/LLM/OVD steps, along with the code to reproduce each step. We rely solely on open source models and all third-party code can be run offline with downloaded weights. \Tabref{tab:modelparam} summarizes the open source models used in our pipeline and their parameter sizes.

\vspace{-10pt}

\paragraph{Contributions} We make the following contributions:
\begin{itemize}
\item A systematic instance annotation pipeline, featuring a novel methodology of pre-detection VLM-LLM re-captioning that enables multi-OVD usage and captures compositional information between instances;
\item A large-scale, high-quality open dataset synthesized with this pipeline.
\end{itemize}

\begin{table}[h]
    \vspace{-5pt}
    \renewcommand{\arraystretch}{0.95}
    \centering
    \begin{tabular}{c|c|c}
        \hline
        \textbf{Model Type} & \textbf{Model Name} & \textbf{Param Size} \\ 
        \hline
        \multirow{2}{*}{VLM} & InternVL1.5 & 26B \\
          & Qwen2VL & 8B\\
        \hline
        LLM & Llama 3 & 8B \\
        \hline
        \multirow{4}{*}{OVD} & Grounding-DINO & 172M \\
        & YOLO-World & 110M \\
        & OWLv2 & 408M \\
        & OV-DINO & 166M \\
        \hline
    \end{tabular}
    \vspace{-5pt}
    \caption{Open source models involved in the creation of \OurDataset{}.}
    \label{tab:modelparam}
    \vspace{-10pt}
\end{table}

\begin{figure*}[ht]
    \centering
    \includegraphics[width=0.98\linewidth]{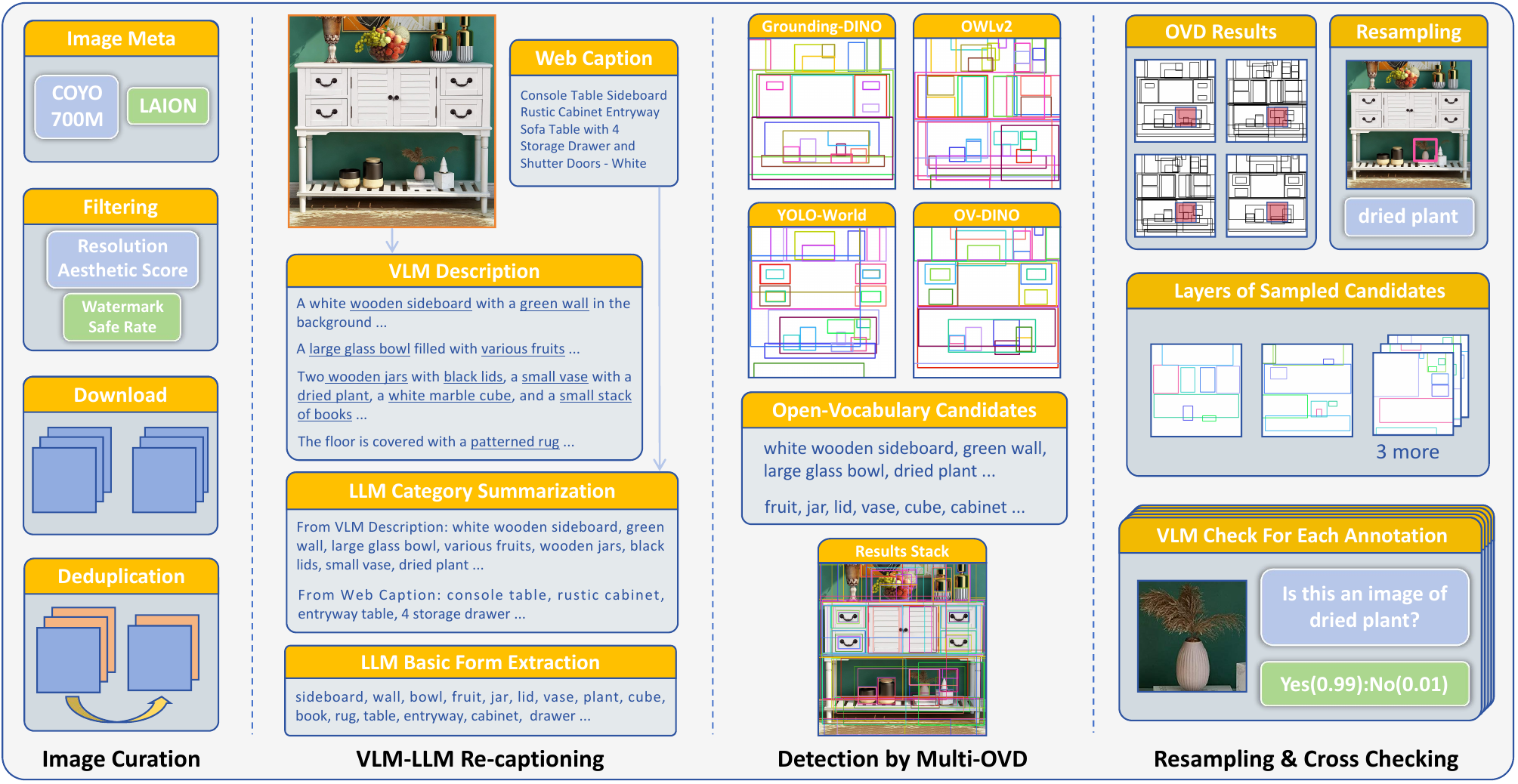}
    \vspace{-6pt}
    \caption{Pipeline overview: stages of our systematic instance annotation framework.
    }
    \label{fig:illustrationPipeline}
    \vspace{-12pt}
\end{figure*}

\vspace{-5pt}
\section{Related work}

\subsection{Text-to-Image Generation}

Text-to-image generation started with GANs~(Generative Adversarial Networks)~\cite{gan}. Recently diffusion~\cite{stablediffusion,sdxl,sd3,dit,pixart,hunyuandit,flux} and auto-regressive~\cite{dalle,parti,llamagen,ARwithoutVQ} methods have achieved a lot of success. By definition, all aforementioned methods take a natural-language text prompt as input and generate an image. Our dataset is specifically designed to support advanced text-to-image generation training, featuring high-quality images paired with comprehensive textual descriptions and others.

\subsection{Language Models and Vision}

With the advent of LLMs~\cite{gpt4,llama2,llama3,qwen2,internlm2,chatglm,deepseek}, language models have received significant attention recently, with numerous approaches integrating them with computer vision tasks. The VLMs~\cite{llava,cogvlm,internvl1.5,qwen2vl,blip2,gpt4v} extend language models with visual capabilities, enabling them to interpret and reason about image content.

In our framework, we use VLMs primarily to generate image descriptions that guide detection and secondarily for cross-checking on detected box-label pairs. While some VLMs can produce bounding boxes directly, they lack the precision and reliability of dedicated detectors. Therefore, we opted to use specialized OVDs for detection purpose.

\vspace{-3pt}
\subsection{Open-Vocabulary Detectors}

Object detection has been a long-standing topic in computer vision. The open-vocabulary variant is flexible in what objects can be detected, although a limited set of potential objects still has to be specified somehow. Among the OVDs (Open-Vocabulary Detectors)~\cite{glip,glipv2,owlv2,ovdino,yoloworld,groundingdino,detclipv3,cvprovd1,cvprovd2} we surveyed, phrase-grounding models~\cite{glip,glipv2,groundingdino,detclipv3} such as Grounding-DINO\cite{groundingdino} take raw text as input and select relevant phrases to associate with each detected bounding box. Other OVDs require a list of potential object categories in text form, with each category potentially containing multiple words.

\vspace{-3pt}
\subsection{Datasets for Detection and Image Description}

Traditional vision datasets such as COCO~\cite{COCO}, LVIS~\cite{LVIS}, Object 365~\cite{obj365}, and Open Images~\cite{openimages} have limited categories. V3Det~\cite{v3det} expands the category scope by building a large taxonomy of 13,204 categories, improving the diversity of instance descriptions. However, when repurposed for text-to-image generation, such datasets pose shared challenges including low image quality, terse image descriptions, and the lack of global contextual connections. Their instance labels also exhibit limited attributes, which is essential for capturing nuanced object characteristics.

Recent efforts have explored detailed image descriptions with compositional information, also an important component for enhancing text-to-image generation, though they lack the explicit instance grounding our work addresses. DOCCI~\cite{docci} and ImageInWords~\cite{imageinwords} datasets emphasize detailed, human-annotated descriptions that capture richer object attributes and improved spatial relationships. From Pixels to Prose~\cite{frompixels} leverages open-source VLMs to generate dense captions, effectively expanding the dataset scale. 

\vspace{-3pt}
\subsection{Instance Grounding}

As text alone has limited capability to specify spatial constraints such as position and layout, many \emph{instance grounding} methods~\cite{GLIGEN,InstanceDiffusion,MIGC,MIGCpp,ifadapter,cvpr_ground1,trainingfree1,trainingfree2} have been developed to support additional forms of input for image generation, in which bounding boxes and per-instance captions are used to provide precise spatial and semantic guidance. These methods typically extend pretrained text-to-image generators by incorporating such spatial information, while still relying on the original text prompt for global context. 

A key challenge in this paradigm lies in preparing high-quality training data that integrates bounding boxes, object-level tags, and complementary text prompts in a cohesive manner. Several strategies have been proposed in recent algorithm research. GLIGEN~\cite{GLIGEN} directly apply phrase-grounding OVDs~\cite{glip,glipv2,groundingdino,detclipv3} on sourced image captions to establish instance-level grounding. 
MIGC~\cite{MIGC} and MIGC++~\cite{MIGCpp} refined this process by extracting instance descriptions from source captions using a natural language processing tool~\cite{stanza}, thereby improving the attribute binding of OVD-detected tags. 
InstanceDiffusion~\cite{InstanceDiffusion} and IFAdapter~\cite{ifadapter} adopt a more structured approach, using RAM~\cite{RAM} to generate instance annotations from the image instead of its caption, which are then passed through OVDs. A VLM is finally applied to produce captions for independently detected instances. 

To the best of our knowledge, our method is the first to focus on the \emph{pre-detection} stage, using a meticulous combination of VLMs and LLMs to improve the OVD input and supplement the results with compositional information. We also employed non-phrase-grounding OVDs~\cite{yoloworld,owlv2,ovdino} for improve detection recall, which would have been impossible in previous methods where the detector input is unstructured natural text.

\vspace{-3pt}
\section{Method}

\Figref{fig:illustrationPipeline} is an overview of our pipeline. We start from a set of curated images fetched from public datasets. We then describe each image with a VLM and prepare a flat list of categories with an LLM. Categories are detected by OVDs. Finally, we merge and resample the detection results and cross-check detected bounding boxes with a different VLM.

Compared to \textit{ad hoc} dataset synthesis, the main innovation we make is the pre-detection re-captioning. By invoking a novel VLM-LLM strategy before detection, we are able to create an accurate and comprehensive list of potential objects, which allows us to detect a broader set of instances using the same OVDs. The re-captioning also mitigates any bias present in the text part of source datasets, significantly increasing the overall consistency.

The complete pipeline processing 1.012M images required approximately 1,960 A800 GPU hours, averaging 6.97 seconds per image or 0.29 seconds per instance. The computational cost breakdown by stage is: VLM description (820h), LLM summarization (250h), object detection (440h), and VLM cross-checking (450h).

\subsection{Image Curation}

Since the ultimate purpose of our dataset is image generation, we need some basic guarantee on image quality. We start with the large COYO-700M~\cite{coyo700m}, LAION2B-en-aesthetic~\cite{laionaes, laion5b} and LAION POP~\cite{laionpop} datasets, and filter for images with aesthetic score~\cite{laion5b} larger or equal to 5.75 and resolution above $1024\times1024$. We also filter out over-sized images with more than 6144 pixels along the longer edge or 4096 pixels along the shorter edge. After pHash deduplication with a Hamming distance of 10, the curation process yielded a total of 1M image-text pairs out of the sourced billions. Only 4.7\% images can pass our resolution standards, in which only 2.1\% can pass our aesthetic score criteria. The deduplication eliminates 31\% of redundant images. 

\begin{figure}[ht]
    \centering
    \includegraphics[width=1\linewidth]{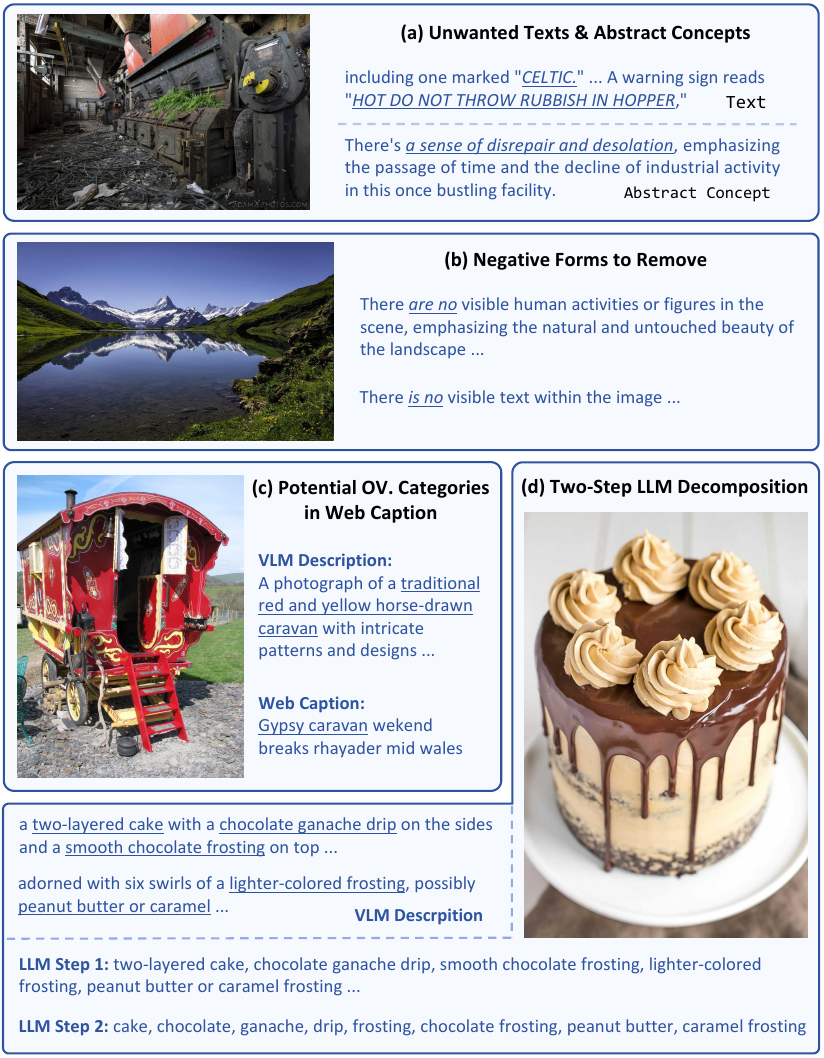}
    \vspace{-12pt}
    \caption{
    Challenges identified in VLM/LLM inference trials. (a)-(b) show possible problematic elements in raw VLM outputs, (c) demonstrates potential open-vocabulary categories in web captions, and (d) illustrates our two-step LLM approach converting detailed VLM descriptions into structured object categories.
    }
    \label{fig:promptwork}
    \vspace{-12pt}
\end{figure}

\subsection{VLM Description}\label{sec:vlmcaption}

As introduced in \secref{sec:intro}, text in source datasets are often terse webpage captions, while we prefer a comprehensive inventory of visual contents. As a solution, we prompt InternVL1.5~\cite{internvl1.5}, a recent VLM, to generate detailed visual descriptions from images. For clarity, we refer to the original dataset text as the \emph{web caption}, and to our generated description as the \emph{VLM description}. 

To accommodate token limits of downstream models, we begin our descriptions with ``This is" followed by the overall image style like ``black and white photograph", preserving such essential information in naïve token clamping. We remove ``This is" as a post-process.

As illustrated in \figref{fig:promptwork}~(a), VLMs might spend tokens discussing non-visual concepts, such as expressions like ``a sense of disrepair and desolation." Similarly, VLMs often transcribe text within images, which can introduce misleading categories for detectors (e.g., ``RUBBISH"). Since such content is difficult to filter otherwise, we dedicate approximately half of our input tokens to explicitly instruct the VLM to avoid such speculation and text transcription.

Additionally, we post-process the VLM output to remove negative statements. As illustrated in \figref{fig:promptwork}~(b), sentences like ``There are no visible human activities" provide little value to detectors and waste resources in subsequent steps.

\vspace{-1pt}
\subsection{LLM Summarization}

Before object detection, we choose to generate a flat list of categories suitable for less flexible OVDs. This text summarization task requires no image processing, so we employ an LLM, Llama3~\cite{llama3}, for its advanced linguistic capabilities. We delay the introduction of the web caption to this step.

As illustrated in \figref{fig:promptwork}~(c), the web caption can contain vital information like ``Gypsy caravan" which cannot be easily derived from visual features alone. However, providing them prematurely in the VLM step risks suppressing the factual ``red and yellow horse-drawn" qualifiers of the same caravan which are equally important. Deferring the web caption to the image-oblivious LLM summarization ensures both versions have a chance to reach downstream OVDs, where anything incorrectly summarized linguistically will likely be ignored.

As shown in \figref{fig:promptwork}~(d), we first instruct the LLM to extract a list of concise categories from the combined text. In this pass, our LLM has a strong tendency to preserve compound phrases in the VLM description, sometimes refusing to break conjunctions like ``peanut butter or caramel frosting". While such behavior enhances attribute binding, the lack of simple words can challenge downstream detectors and limit our coverage of simple constituent objects.

As mitigation, we implement a second decomposition step to extract basic forms, like standalone nouns, from the output phrases. Specifically, we prompt the LLM to break complex concepts into their constituent parts (e.g., extracting \emph{cake}, \emph{chocolate}, \emph{ganache}, and \emph{drip}). We then combine the results from both steps as OVD input. The redundancy improves recall by providing core components as backup when a complex phrase fails detection.

\vspace{-1pt}
\subsection{Object Detection}\label{sec:objectdetection}

For the actual detection, we employ Grounding-DINO~\cite{groundingdino}, OV-DINO~\cite{ovdino}, YOLO-World~\cite{yoloworld} and OWLv2~\cite{owlv2}, sending the category list to all four OVDs. For each OVD, we set its confidence score threshold slightly lower than conventional usage to retain more candidate boxes while balancing their average detection counts. We concatenate the boxes from all OVDs and perform IoU~(Intersection over Union)-based NMS~(Non-Maximum Suppression) with a threshold of 0.4 between boxes of the same category.

After concatenation, we apply a resampling process to address excessive box coverage resulted from using multiple OVDs. This process balances OVDs, penalizes overlapping boxes, duplicate captions, distance from image center, and small box sizes, effectively removing approximately 70\% of instances while preserving detection quality. Further details on this resampling methodology are provided in the supplementary materials.

\Tabref{tab:multiovd} presents the distribution of detection boxes and category coverage across our four OVDs on 5K randomly sampled images. Each OVD demonstrates distinct strengths in different categories, and a considerable portion of categories can only be detected by one of them. Such multi-detector fusion is enabled by our fine-grained VLM description and LLM summarization, which produces the flat category list required by the less flexible OVDs~\cite{yoloworld,owlv2,ovdino}.

\begin{table}[h]
\vspace{-5pt}
  \setlength{\tabcolsep}{3.5pt}
  \renewcommand{\arraystretch}{1.00}
  \centering
  \begin{tabular}{c|cccc} 
    \hline
    \textbf{OVD Type} & GD~\cite{groundingdino} & YW~\cite{yoloworld} & OW~\cite{owlv2} & OD~\cite{ovdino} \\
    \hline
    Box Contribution & 26.7\% & 21.7\% & 28.0\% & 23.6\% \\
    \hline
    Cat. Coverage & 49.5\% & 42.9\% & 49.1\% & 31.2\% \\
    Unique Cat. & 17.0\% & 11.9\% & 15.8\% & 9.4\% \\
    \hline
  \end{tabular}
  \vspace{-5pt}
  \caption{Per-OVD statistics on 5K images. GD: Grounding-DINO, YW: YOLO-World, OW: OWLv2, OD: OV-DINO.}
  \label{tab:multiovd}
\vspace{-5pt}
\end{table}

\vspace{-5pt}

\subsection{VLM Cross-Checking}\label{sec:vlmcheck}

At the end of our pipeline, we crop out the bounding box of each resampled instance. We subsequently prompt a second VLM, Qwen2VL~\cite{qwen2vl}, to check whether the image content matches its expected caption. This process can be expensive due to running a VLM or each box. We manage the cost by asking a yes/no question and restricting the output token count to 1. We only keep an instance when the total probability of all capitalization variants of ``yes" is significantly more than that of ``no".

\begin{figure*}[ht]
    \centering
    \vspace{-2pt}
    \includegraphics[width=0.97\linewidth]{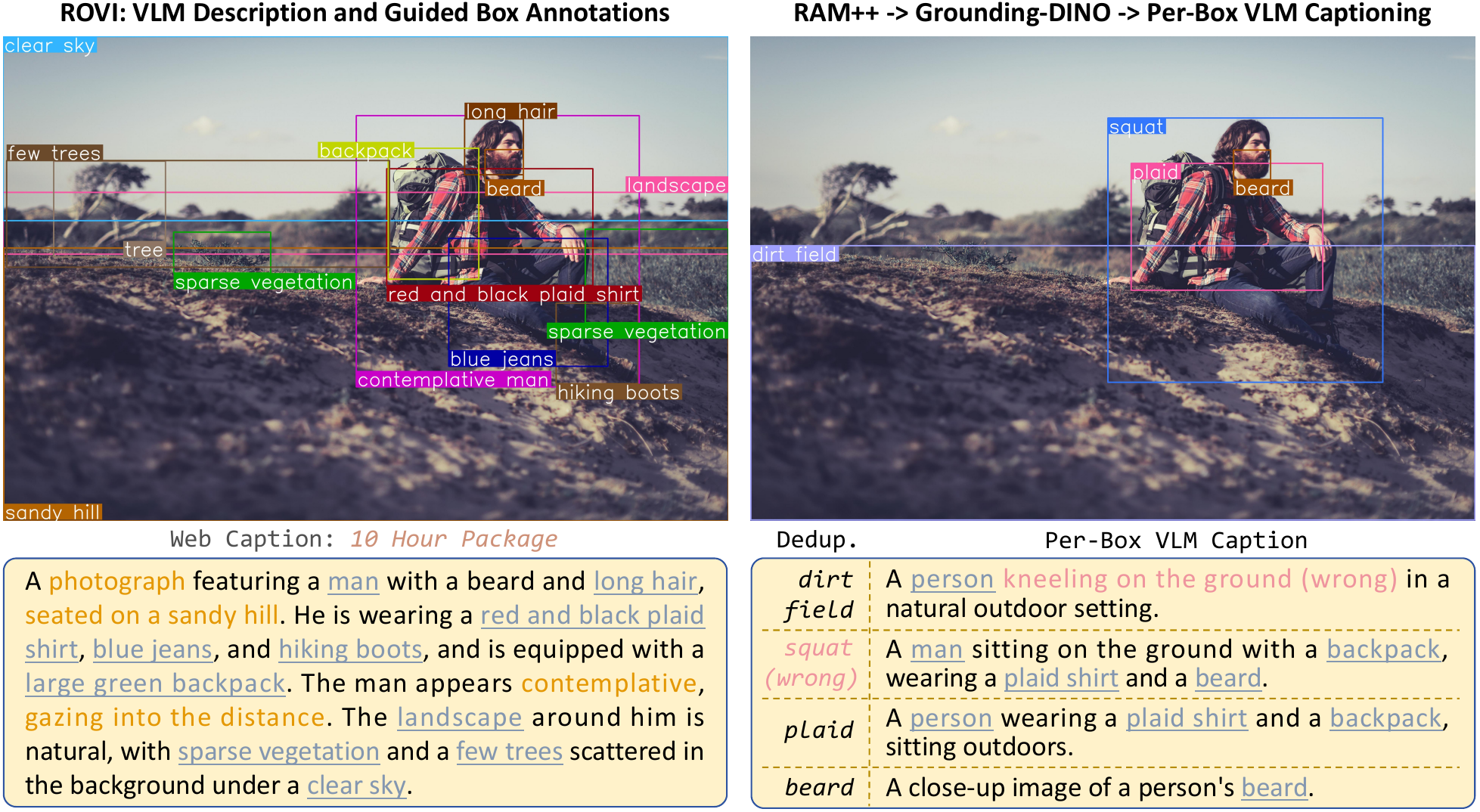}
    \vspace{-7pt}
    \caption{Effect of RAM++ tagging plus per-box VLM captioning. RAM++ tags: \emph{dirt field, floor, hill, hillside, land, mound, terrain, squat, man, sit, plaid, shirt, beard}. Note the high redundancy within this list. For clarity, detected boxes are de-duplicated in both approaches.
    }
    \label{fig:figforRAM}
    \vspace{-13pt}
\end{figure*}

\section{Experiments}

\subsection{Study on Data Generation Alternatives}\label{sec:gencategory}

Our primary contribution is the pre-detection VLM-LLM re-captioning. In this section, we assess its impact by comparing it with alternatives that served similar roles in previous instance grounding works.

The first group relies on recent phrase-grounding OVDs, such as GLIP, GLIPv2, and Grounding-DINO, to extract categories as token subsets from natural language text. Among those OVDs, Grounding-DINO offers state-of-the-art performance, which we choose as a comparison target following established practice~\cite{MIGC, MIGCpp, InstanceDiffusion, ifadapter}. We tested two variants of this straightforward approach:

\begin{itemize}
    \item \textbf{Web Cap. + OVD} uses the web captions directly as input to phrase-grounding OVD, similar to the approach adopted by GLIGEN~\cite{GLIGEN}.
    \item \textbf{VLM Desc. + OVD} follows \secref{sec:vlmcaption} to generate a VLM description of the image, which is sent to the OVD. This is a straightforward extension of the GLIGEN approach to VLMs.
\end{itemize}

The second group follows methods first introduced in InstanceDiffusion~\cite{InstanceDiffusion}, which operates in two stages: first performing recognition-based tagging on images and inputting these tags to OVD for detection, then generating VLM captions individually for each detected box, producing isolated instance-level descriptions. For the tagging component, we tested both RAM~\cite{RAM} and RAM++~\cite{RAM}.

\Tabref{tab:vsdetgencat} compares the number of categories and bounding boxes extracted using each approach on 5K randomly sampled images. RAM tags allowed the detectors to discover more boxes but are limited by predefined categories. The OVD-extracted categories are more diverse, while detecting less boxes. Our dataset outperforms both significantly.

\Figref{fig:figforRAM} uses a concrete example to illustrate the difference between our approach and RAM-based alternatives. While RAM tagging addresses the web caption bias, many less prominent objects remain undetected. They also tend to produce boxes for non-object concepts such as ``squat" or ``plaid," which can be counterproductive for image generation.

The right figure also illustrates the unexpected effect of the per-box VLM captioning strategy. While non-object boxes receive more reasonable captions, all boxes overlapping the person described him to some extent. Such content duplication across differently shaped boxes can mislead the trained image generator. Furthermore, the VLM hallucinated details based on incomplete crops, such as interpreting a cropped human body showing only legs as ``kneeling on the ground." In contrast, the pre-detection re-captioning allows \OurDataset{} to provide higher object detection coverage, better context integration, and improved category precision.

\begin{table}[h]
    \vspace{4pt}
    \centering
    \setlength{\tabcolsep}{5pt}
    \renewcommand{\arraystretch}{0.99} 
    \begin{tabular}{c|ccc}
        \hline
        \textbf{Data Gen. Method} & \textbf{Cat.} & \textbf{Avg. Cat.} & \textbf{Avg. Box}\\ 
        \hline
        Web Cap.$\to$OVD & 3001 & 0.72 & 0.86 \\
        VLM Desc.$\to$OVD & 14412 & 7.67 & 9.20 \\
        \hline
        RAM$\to$OVD & 1836 & 6.44 & 11.54 \\
        RAM++$\to$OVD & 1846 & 6.49 & 11.58 \\
        \hline
        \OurDataset{}'s & \textbf{25374} & \textbf{12.45} & \textbf{24.01} \\
        \hline
    \end{tabular}
    \vspace{-6pt}
    \caption{Quantitative comparisons between data generation methods. RAM/RAM++ settings follow the InstanceDiffusion~\cite{InstanceDiffusion} implementation. We only count box-caption pairs that passed our VLM cross-check in \secref{sec:vlmcheck}. \emph{Cat.}: number of distinct categories discovered over the entire sampled set. \emph{Avg.}: per-image averages. }
    \label{tab:vsdetgencat}
    \vspace{-4pt}
\end{table}

\begin{table}[ht]
  \setlength{\tabcolsep}{3.3pt}
  \renewcommand{\arraystretch}{0.98}
  \centering
  \begin{tabular}{c|cccc} 
    \hline
    \textbf{Model} & \textbf{Gen Inst.$\uparrow$} & \textbf{FID$\downarrow$} & \textbf{Aes.$\uparrow$} & \textbf{CLIP Sim.$\uparrow$} \\
    \hline
    SD v1.4 & \compare{0.530}{0.398} & \compare{17.6}{38.3} & \compare{5.66}{5.17} & \compare[,bold]{0.283}{0.257} \\
    Offi. G. & \compare{0.750}{0.812} & \compare{19.3}{25.4} & \compare{5.33}{4.78} & \compare{0.273}{0.233} \\
    MIGC & \compare{0.821}{0.833} & \compare{27.6}{32.2} & \compare{5.47}{5.23} & \compare{0.228}{0.213} \\
    Instdiff & \compare{0.854}{0.871} & \compare{21.1}{28.4} & \compare{5.47}{5.22} & \compare{0.262}{0.229} \\
    \hline
    \textbf{Our G.} & \compare[bold,bold]{0.880}{0.872} & \compare[bold,bold]{15.7}{16.7} & \compare[bold,bold]{5.83}{5.40} & \compare[bold,]{0.286}{0.245} \\
    
    \hline
  \end{tabular}
  \vspace{-6pt}
  \caption{Apples-to-apples quantitative comparisons on ROVI validation set (left values) and Open Images validation set (right values in brown). \emph{Gen Inst.}: per-box cross-checking pass rate using VLM-based evaluation. \emph{Our G.}: GLIGEN trained on \OurDataset{}. \emph{SD v1.4}: Stable Diffusion baseline~\cite{sdv1.4}. \emph{Offi. G.}: Official GLIGEN~\cite{GLIGEN}. \emph{InstDiff}: InstanceDiffusion~\cite{InstanceDiffusion}. \emph{Aes.}: Aesthetic score~\cite{laionaes}. \emph{CLIP Sim.}: CLIP-based image-text similarity~\cite{clip}.}
  \label{tab:rovigenstats}
\vspace{-11pt}
\end{table}

\vspace{-2pt}
\subsection{Demonstrations on Generated Images}

To evaluate the end-to-end effect of our dataset, we trained a GLIGEN network on the training portion of our dataset, using the official Stable Diffusion v1.4 checkpoint as the base model. This configuration is deliberately chosen to better isolate improvements induced by our dataset from algorithmic improvements in newer methods. During comparison we generate images using each method from the same text prompt and the same set of box-label pairs.

\vspace{-5pt}
\paragraph{Quantitative Results.}

\Tabref{tab:rovigenstats} compares qualitative metrics between our GLIGEN, existing methods, and a Stable Diffusion baseline. We stochastically sampled 5000 prompt-image pairs from both the ROVI validation set (black numbers) and the Open Images validation set~\cite{openimages} (brown numbers) and compared instance-level and image-level metrics on generated images.

At the instance level, we focus on instance grounding accuracy, evaluated using the VLM-based cross-checking method in \secref{sec:vlmcheck}. Specifically, we cropped each expected instance from its specified bounding box on a generated image and used Qwen2VL~\cite{qwen2vl} to compute the likelihood of the cropped window matching its assigned label. Our GLIGEN has achieved pass rates comparable with more recent methods on both datasets.

At the image level, our GLIGEN consistently outperforms existing alternatives quality-wise (\emph{FID, Aes.}) while also remaining more faithful to the global prompt (\emph{CLIP Sim.}). Notably, we also outperformed the baseline quality-wise, even on the Open Images validation set. Such image-level metrics are known to disadvantage fine-tuned models as noted in the MIGC~\cite{MIGC} paper, and the results indicate that training GLIGEN on ROVI improved image quality sufficiently to offset this disadvantage. We attribute this to our improved grounding and prompt quality.

Comparing the two validation datasets, the Open Images set results in lower image quality for every method. Our manual inspection observed visually objectionable artifacts in a large portion of generated images regardless of model. We hypothesize that the degraded performance stems from three properties critical for grounded text-to-image evaluation: (1) comprehensive open-vocabulary labels, (2) high-quality captions suitable as prompts, and (3) aesthetic quality matching generation standards. For example, the Open Images dataset we used only contains 600 categories. Its image captions are seldom consistent with image generation prompts. 
Many images in the set are also aesthetically questionable for evaluation purposes. 
We have explored commonly used instance-annotated datasets and found they all have similar problems, as they were mostly designed for detection tasks. In contrast, our ROVI validation set was specifically curated to address these limitations, which can provide more reliable evaluation for real-world grounded text-to-image generation scenarios.


\begin{figure*}[ht]
\vspace{-5pt}
    \centering
    \includegraphics[width=0.98\linewidth]{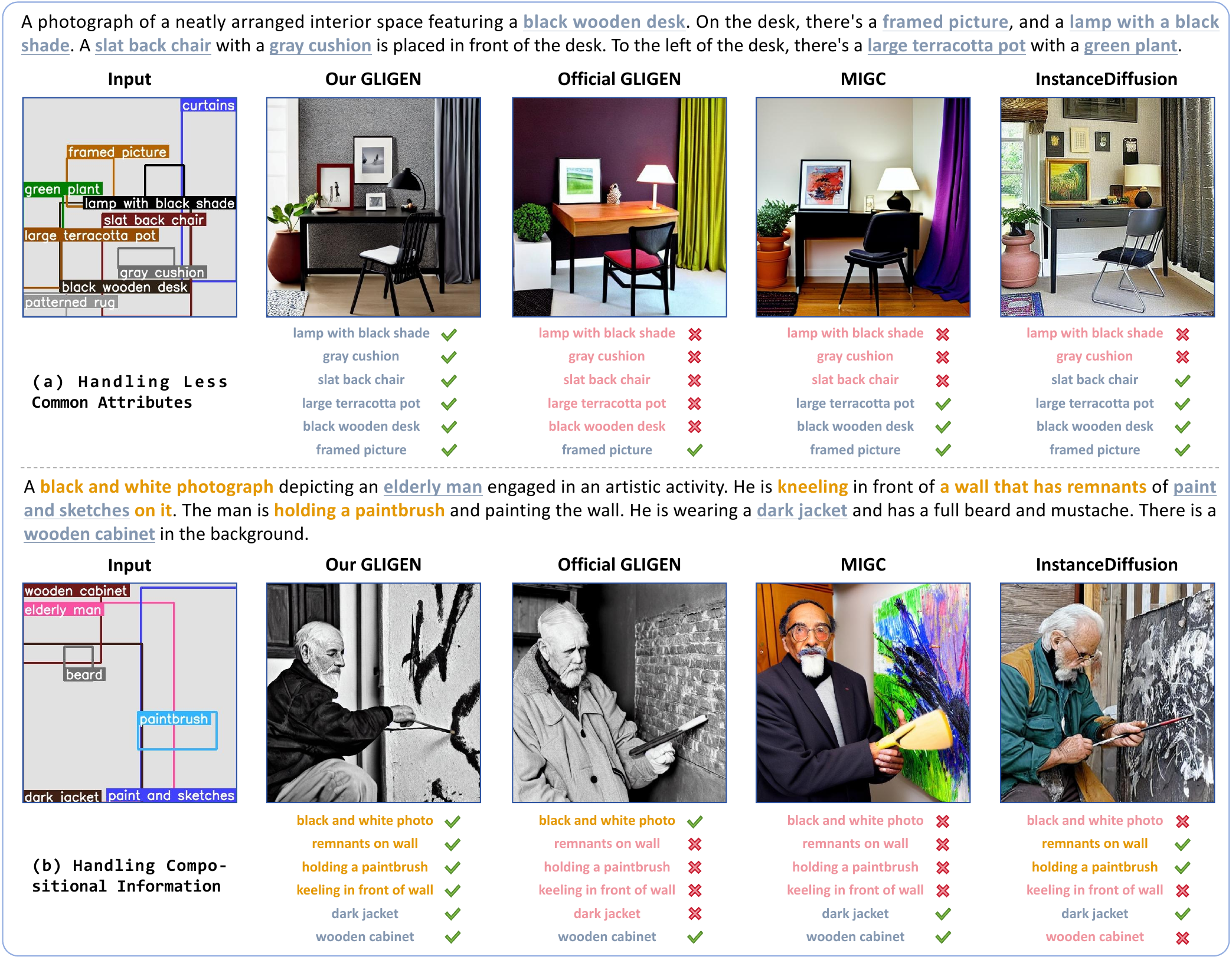}
    \vspace{-6pt}
    \caption{Comparing model performance on compositional prompts with instance grounding. Besides, note the attribute binding issues such as \emph{gray} cushion. Grayish-blue: instance attributes. Brown: compositional elements. Red: mismatch in generation.
    }
    \label{fig:qualitative}
    \vspace{-14pt}
\end{figure*}

\vspace{-4pt}
\paragraph{Qualitative Results.}

We also present qualitative results to evaluate attribute bindings and object composition. To make the results more interpretable for visualization and user study, we hand-picked box-label pairs that provide a semantically meaningful description of the global scene.

\Figref{fig:qualitative}~(a) compares the reproduction accuracy of instance attributes. All methods successfully produced the correct object types such as ``lamp" or ``chair" at the requested locations. However, when both the prompt and instance grounding specified a ``lamp with \emph{black shade}," only our model trained on \OurDataset{} correctly rendered the \emph{black shade}, while other algorithms defaulted to \emph{white shades}. Similar accuracy improvements were observed with the ``slat back chair".

We attribute this improved attribute fidelity to the category diversity enabled by our VLM-LLM re-captioning strategy. While visually evident, attributes like lamp shade colors and chair back types are rarely mentioned in conventional image-text pairs. We direct the VLM to comprehensively catalog these self-evident attributes, which are then fully revealed by the LLM, allowing subsequent OVDs to detect and ultimately enabling their accurate generation.

\Figref{fig:qualitative}~(b) illustrates how the compositional information in VLM descriptions improves text-image alignment. When prompted to generate \emph{a black and white photograph of an elderly man kneeling in front of a wall and holding a paintbrush}, networks trained on other datasets produced correct overall layouts but neglected compositional details, suggesting over-reliance on instance grounding. In contrast, our GLIGEN faithfully reproduced most details, a capability directly enabled by our re-captioning methodology.

\vspace{-4pt}
\paragraph{User Study.} We have also conducted a user study to analyze how well our trained GLIGEN network respects instance grounding, the prompt, and its aesthetic quality when compared to alternatives. We ran all candidate implementations to synthesize images from 200 box-grounded inputs. Each survey session presented two images from the same input alongside their shared instance grounding and text prompt, and asked the user to select which image they prefer aesthetically / better aligns with the prompt. We also asked the users to specify which image better respects the grounding constraint of each box-label pair. The image pairs had their placement randomized so that users could not infer their sources. The users are volunteered researchers with ample expertise in image generation.

\begin{table}[h]
\vspace{-5pt}
    \centering
    \setlength{\tabcolsep}{5pt}
    \renewcommand{\arraystretch}{1.00} 
    \begin{tabular}{c|ccc}
        \hline
        \multirow{2}{*}{\textbf{Our GLIGEN vs}} & \multicolumn{3}{c}{\textbf{Win Rate}} \\
        \cline{2-4}
         & Official G. & MIGC & InstDiff\\ 
        \hline
        Instance Alignment & 82.0\% & 94.8\% & 85.0\% \\
        Prompt Alignment  & 83.3\% & 90.5\% & 82.3\% \\
        Aesthetic Quality  & 85.3\% & 98.3\% & 83.3\% \\
        \hline
    \end{tabular}
    \vspace{-5pt}
    \caption{Ratio of users prefering GLIGEN~\cite{GLIGEN} trained on \OurDataset{} compared to each alternative on each aspect. \emph{Official G.}: official implementation of GLIGEN. \emph{InstDiff}: InstanceDiffusion~\cite{InstanceDiffusion}.}
    \label{tab:humaneval}
    \vspace{-9pt}
\end{table}

\Tabref{tab:humaneval} shows the result of the user study. GLIGEN trained on \OurDataset{} is preferred in all aspects by a decisive margin. \OurDataset{} allows the network to learn many subtly different attributes of each core object, thanks to our more diverse categories and more detected boxes. The improved alignment appears obvious to human users and significantly contributes to the overall quality of the final images.


\vspace{-5pt}
\section{Limitations and Discussion}
\vspace{-4pt}

We have presented \OurDataset{}, a synthetic dataset for instance-grounded text-to-image generation that addresses key limitations in existing approaches. By employing VLMs for comprehensive image description followed by LLM-based open-vocabulary category summarization, our pre-detection strategy enables more complete instance grounding and superior text-to-image generation.

We manually inspected 24k bounding boxes from 1,000 randomly selected images, finding that 3.3\% of the detected boxes contain errors. These errors stem from OVD false-positives that bypassed VLM cross-verification. The two main failure modes are: (1) detecting a partially occluded object as several small visible parts, and (2) misinterpreting phrases, such as labeling a rock as a ``book about rocks".

A few minor language model artifacts persist through our pipeline, including inconsistent singular/plural handling, left/right confusion when models interpret human figures as mirrored, and awkward phrasing such as ``blue dressed woman" or ``stethoscope-wearing woman". Bounding box grounding is also inherently inaccurate for visually occluded objects and non-contiguous elements. Our resampling strategy partially mitigates these inaccuracies.

While our dataset features high-quality, high-resolution images, current GLIGEN based on Stable Diffusion v1 experiments cannot fully utilize these improvements. Newer architectures would better showcase \OurDataset{}'s strengths in high-resolution rendering, aesthetic quality, and complex grounding. We aim to implement our dataset with advanced models like Flux-dev~\cite{flux} as future work.

We refer readers to our supplementary materials for more detailed examples and limitation discussions.

\vspace{10pt}

\makeappendix

This appendix is structured into several sections that provide additional details, aiming to present more cases and analyses in image illustrations. Specifically, it will cover the following topics:


\begin{itemize}
\item \secref{sec:generation}: Additional generation comparisons between our GLIGEN~\cite{GLIGEN} model trained on \OurDataset{} and alternative methods~\cite{stablediffusion,GLIGEN,MIGC,InstanceDiffusion} trained on different datasets.
\item \secref{sec:dataset}: Dataset snapshots and analysis demonstrating \OurDataset{}'s superior quality and diversity advantages through representative examples across various visual domains.
\item \secref{sec:resample}: Detailed illustrations and explanations of our resampling stage methodology.
\item \secref{sec:limitation}: Cases illustrating the limitations of \OurDataset{}'s annotations.
\item \secref{sec:failure}: Analysis of failure cases in our GLIGEN model's generation process.
\item \secref{sec:userstudy}: Additional details regarding the user study presented in the main paper.
\end{itemize}

Please note that we have limited the display of bounding box annotations. \figref{fig:limitation_woman} presents a fully labeled case; however, the number of bounding boxes exceeds practical limits for clear visualization. Therefore, we selectively annotate only the salient objects in subsequent figures to enhance clarity and readability.

\vspace{1.0em}

\begin{figure}[H]
    \centering
    \includegraphics[width=0.95\linewidth]{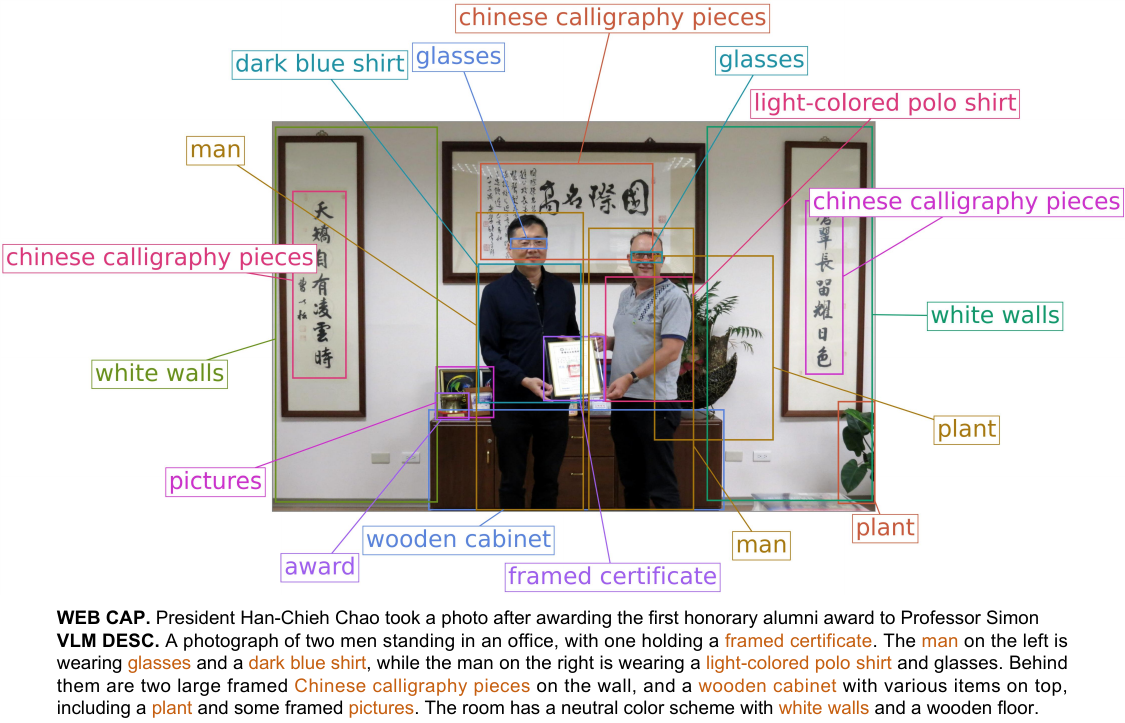}
    \caption{Effect of pre-detection VLM-LLM re-captioning, our key contribution. \emph{Web CAP.} refers to the original text source, while \emph{VLM DESC.} denotes the generated VLM description in \OurDataset{}, specifically utilizing InternVL1.5~\cite{internvl1.5}. In this context, all detected entities are summarized by an LLM, Llama3~\cite{llama3}, from the exhaustive VLM description. The detection effectively describes the image, and the detailed open-vocabulary categories offer a clearer view of the visual content, significantly surpassing traditional detection results that rely on basic categories. Furthermore, instance grounding in image generation can benefit from the richness of categories and compositional elements present within the VLM description.}
    \label{fig:calligraphy}
\end{figure}

\newpage
\setcounter{page}{2}
\section{\centering More Comparisons on Generation}\label{sec:generation}

\begin{figure}[H]
    \centering
    \includegraphics[width=1\linewidth]{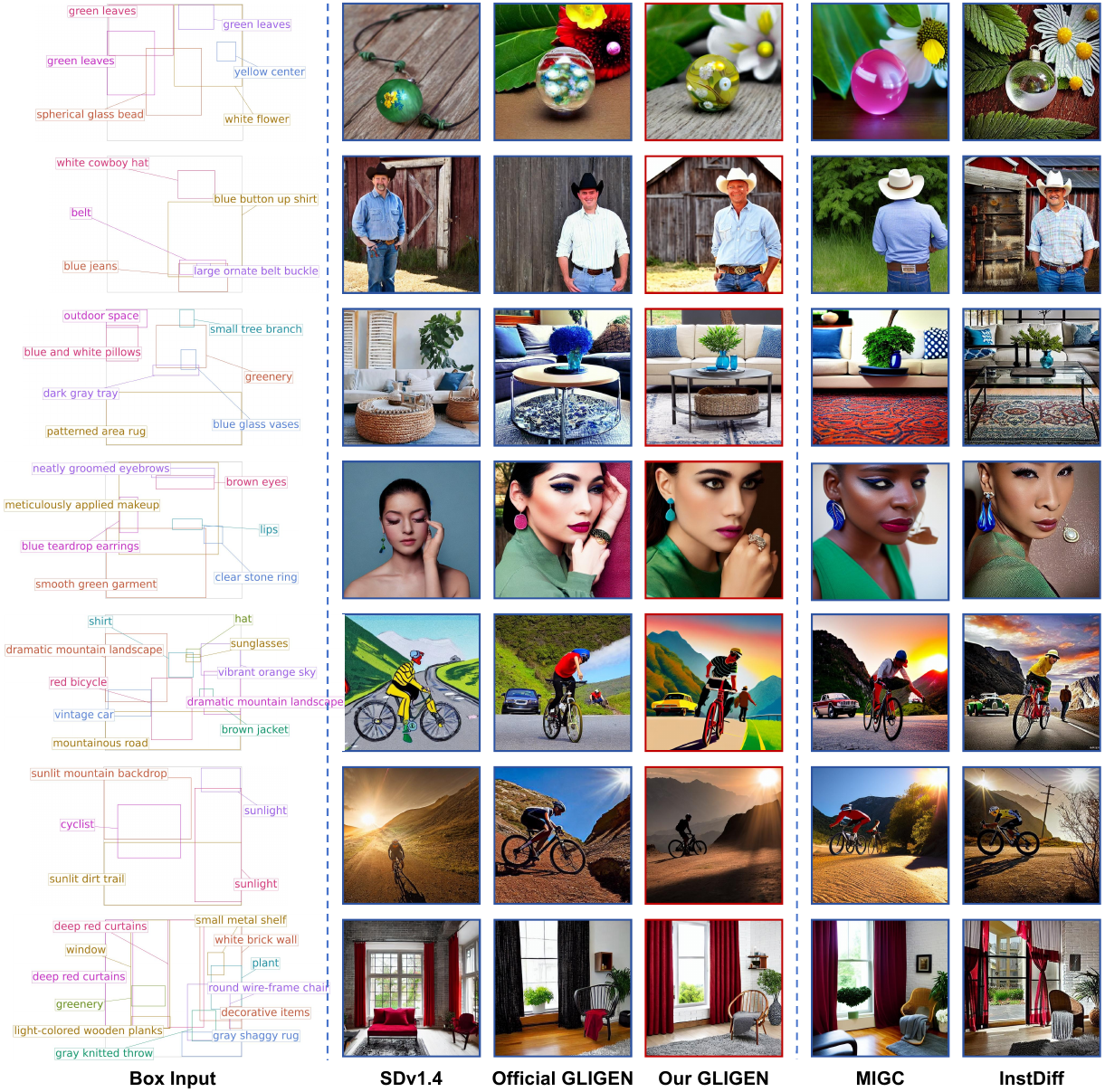}
    \caption{The above presents a comparison of the generated results. Note that the GLIGEN model trained on \OurDataset{} (\emph{Our GLIGEN}) differs from the official GLIGEN~\cite{GLIGEN} trained on other datasets. \emph{InstDiff} refers to InstanceDiffusion~\cite{InstanceDiffusion} All results are based on the test data mentioned in the paper's main text, where both bounding box annotations and prompts have undergone manual verification. We omit prompts for readability. Note the combinations of attributes formed by color, texture, and overall coherence of the images, as well as their aesthetic qualities.}
    \label{fig:infer_cases_1}
\end{figure}

\newpage
\begin{figure}[H]
    \centering
    \includegraphics[width=1\linewidth]{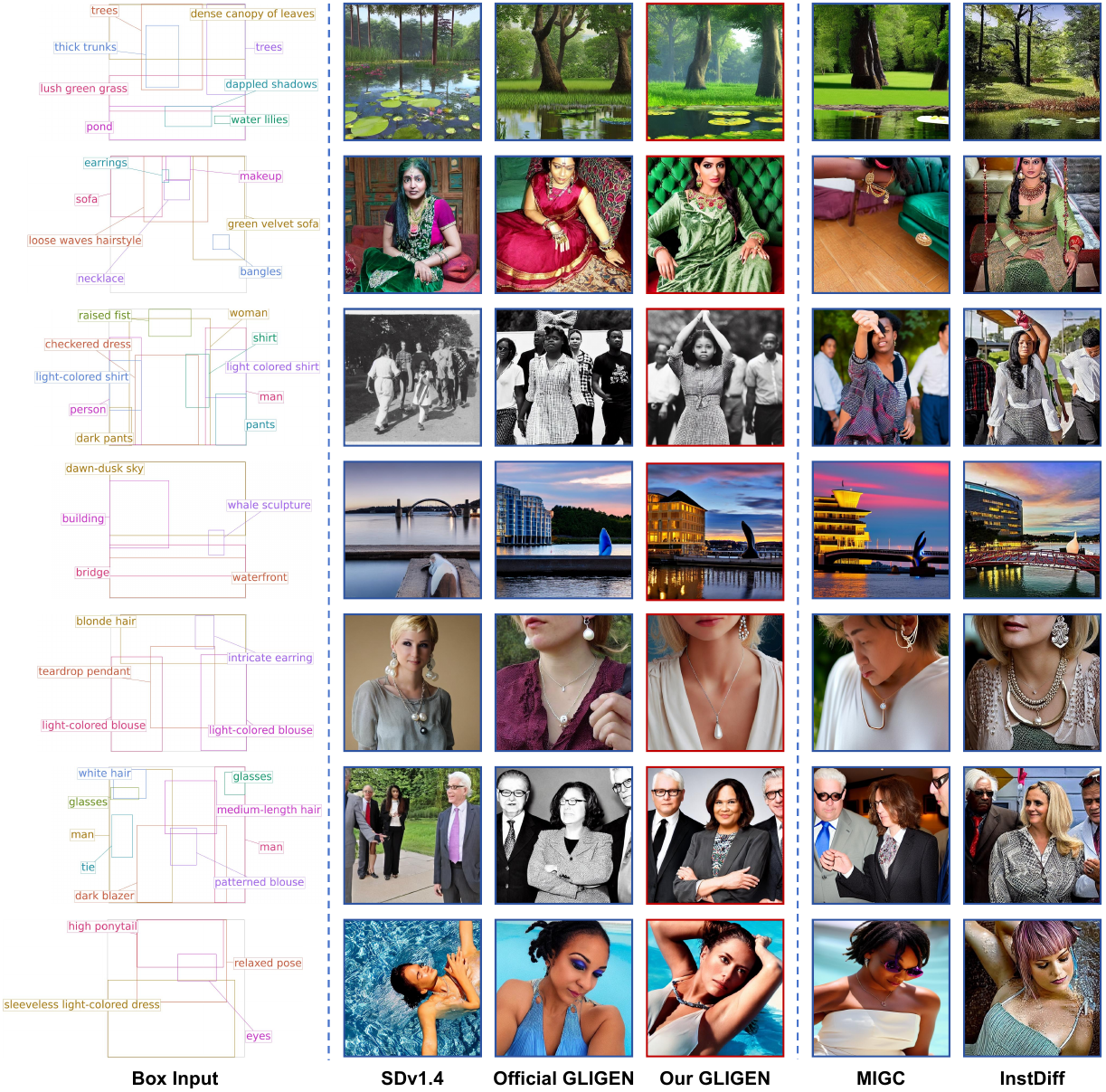}
    \caption{In the results mentioned above, the prompt in the third row requests the generation of a \emph{black-and-white} image; however, neither MIGC~\cite{MIGC} nor InstanceDiffusion achieved this objective. Notably, there are some specific open-vocabulary categories, such as \emph{whale sculpture} in the fourth row, \emph{teardrop pendant} in the fifth row, and \emph{medium-length hair} in the sixth row. If these open-vocabulary categories are not correctly understood, the generated content may deviate from the requirements, resulting in inaccuracies such as depicting a teardrop as a circular shape or rendering medium-length hair as long hair.}
    \label{fig:infer_cases_2}
\end{figure}

\newpage
\begin{figure}[H]
    \centering
    \includegraphics[width=0.82\linewidth]{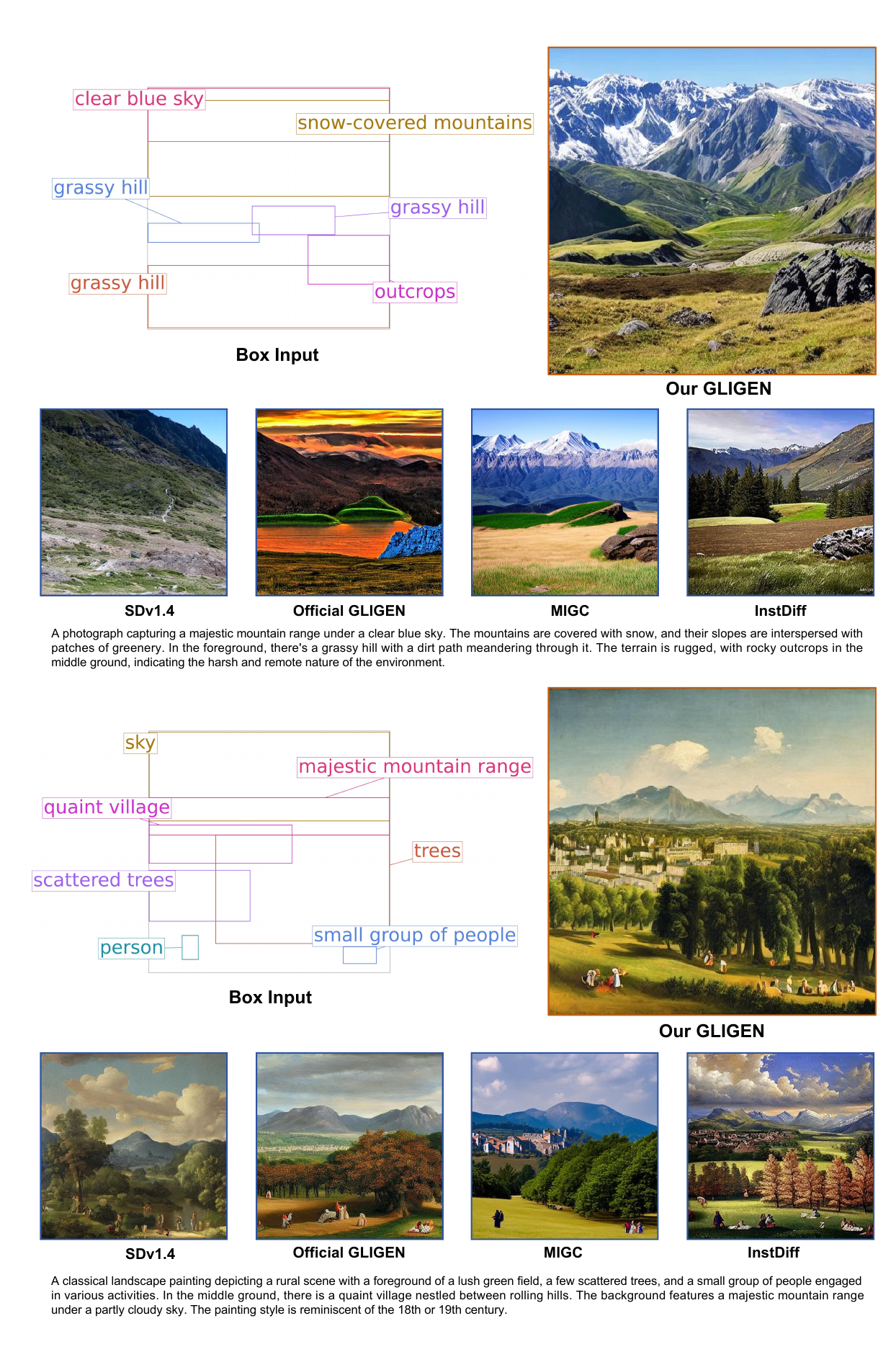}
    \caption{The superior visual quality and coherence demonstrated in our generated images can be attributed to the rigorous curation standards employed in ROVI. Specifically, ROVI's emphasis on high-resolution source images and stringent aesthetic quality criteria ensures that the training data better aligns with the demands of contemporary text-to-image synthesis tasks. This careful dataset construction directly translates to improved generation capabilities, as evidenced by the comparative results shown in the accompanying figures.}
    \label{fig:infer_zoom}
\end{figure}

\newpage

\section{\centering Dataset Snapshots \& Advantages Overview}\label{sec:dataset}

This section provides a comprehensive analysis of the key advantages of the ROVI dataset, followed by representative examples across diverse visual domains that demonstrate these capabilities in practice.

\subsection{Dataset Advantages Overview}

\begin{figure}[H]
    \centering
    \includegraphics[width=0.9\linewidth]{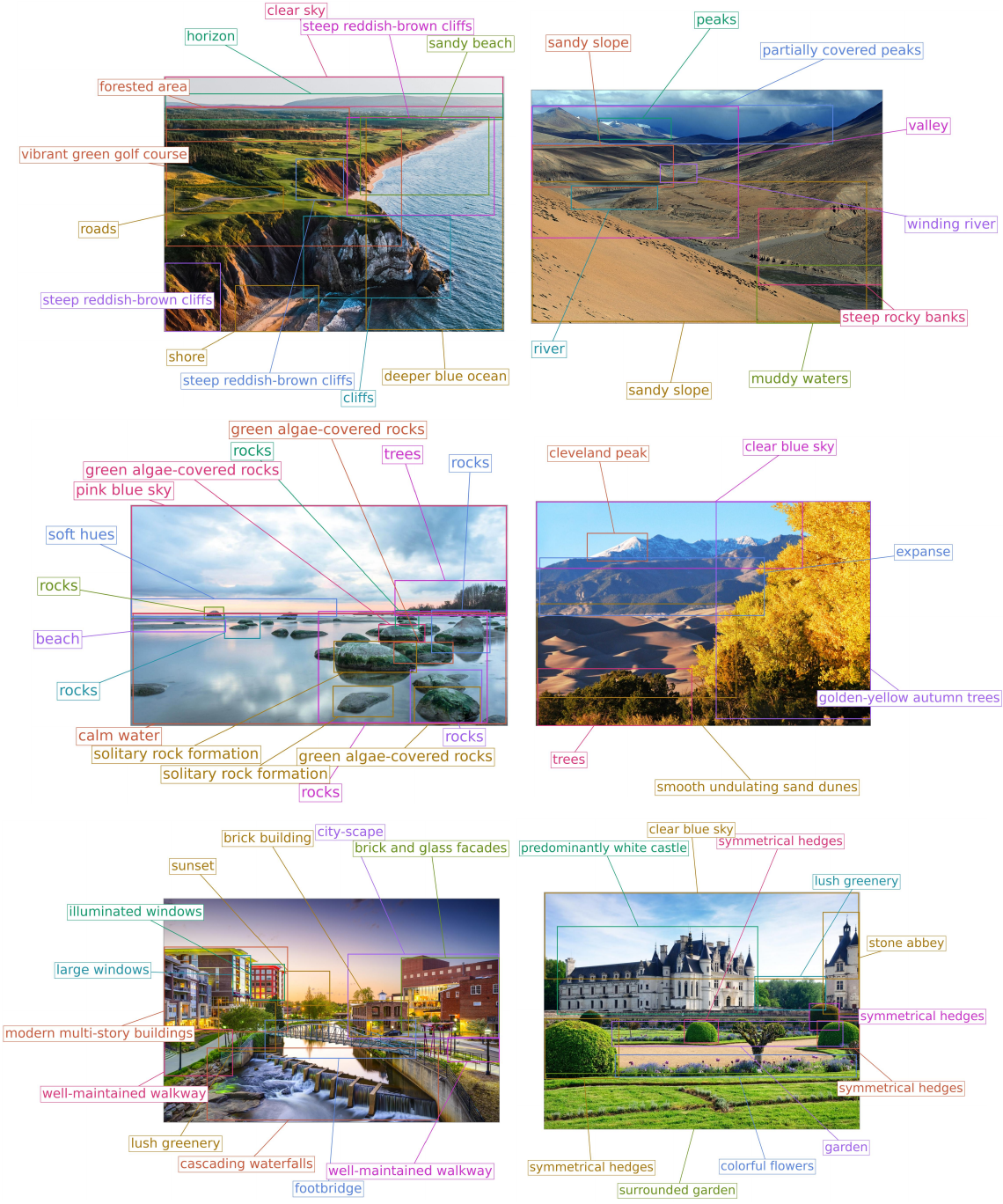}
    \caption{Landscapes constitute a significant proportion of our high-aesthetic dataset. 
    Traditional detection approaches with restricted vocabularies typically produce monotonous 
    summarization with minimal detail. Our dataset provides rich and precise annotations that 
    effectively capture key features across various natural scenes. The middle-right example 
    demonstrates Named Entity Recognition (NER) capabilities, where \emph{cleveland peak} preserves 
    specific geographical information from the original web caption that would be lost in 
    generic detection approaches. Such detailed open-vocabulary categories offer substantially 
    clearer characterization of landscape elements compared to basic detection outputs.}
    \label{fig:landscapes}
\end{figure}

\begin{figure}[H]
    \centering
    \includegraphics[width=0.95\linewidth]{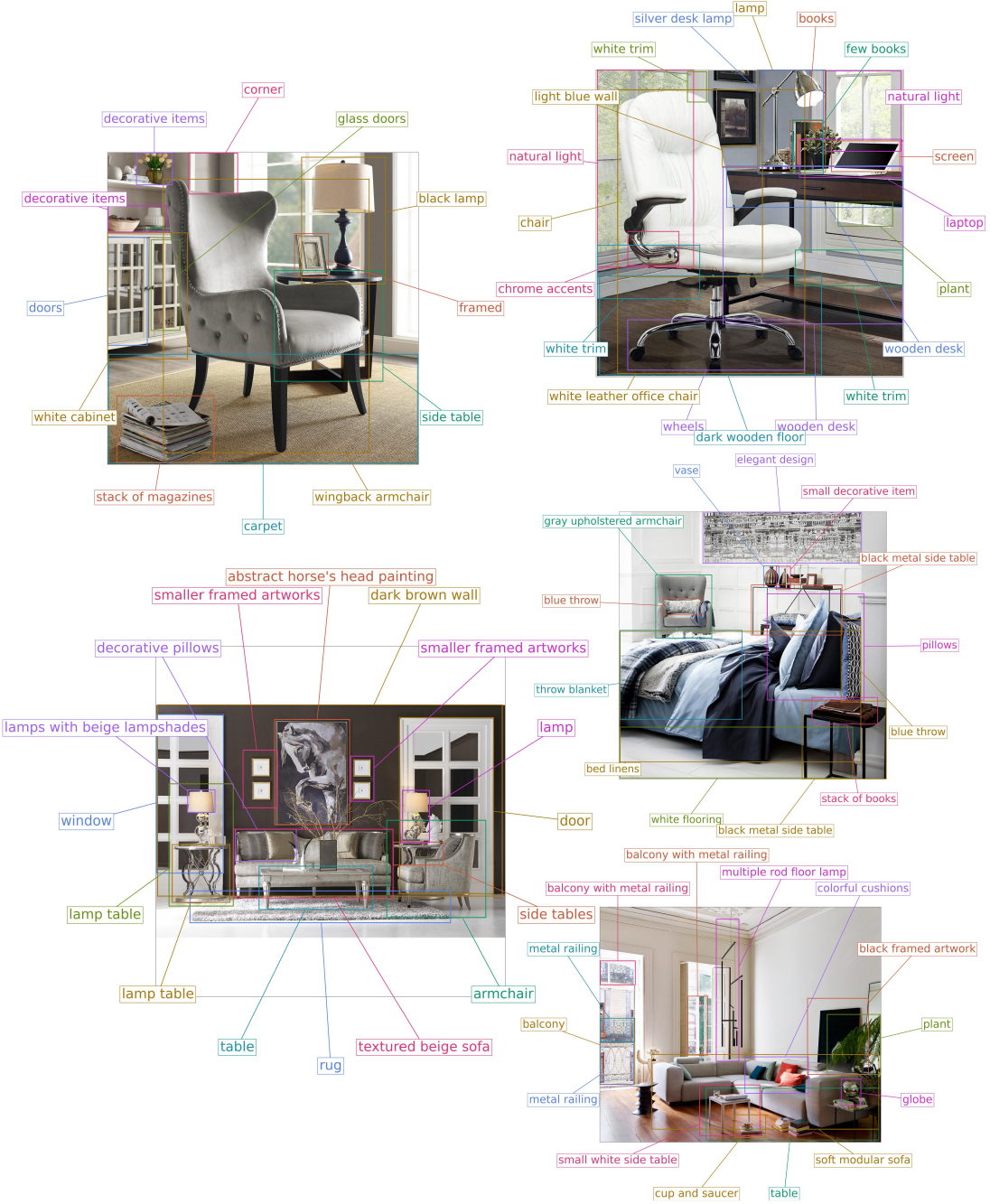}
    \caption{Indoor scenes involve numerous objects and complex visual hierarchies that 
    challenge traditional detection methods. The illustrated examples demonstrate that our 
    VLM-LLM re-captioning workflow achieves comprehensive coverage even for intricate indoor 
    scenarios. Open-vocabulary categories show notable advantages in capturing color variations, 
    texture details, and combinatorial relationships between objects. The examples reveal how 
    our approach maintains contextual understanding of object arrangements and spatial 
    relationships that are essential for accurate scene representation in indoor environments.}
    \label{fig:indoor_setting}
\end{figure}

ROVI is specifically designed to meet the demanding requirements of advanced grounded text-to-image generation, where precise spatial control and rich semantic understanding are essential. Compared to existing detection-centric datasets, ROVI provides several key advantages:

\paragraph{Attribute-rich instance labels:} Instance labels include detailed object attributes such as colors, textures, materials, and semantic properties, integrated with contextual descriptions rather than generic category names. Examples include ``green algae-covered rocks" in \figref{fig:landscapes} (middle-left) and ``white leather office chair" in \figref{fig:indoor_setting}.

\paragraph{Complex scene understanding:} Comprehensive coverage of visual hierarchies and relational structures enables detailed characterization of objects, environments, and their compositional interactions. This is best illustrated by the medalists in \figref{fig:resample_ovd}.

\paragraph{Diverse domain coverage:} Broad applicability across varied content domains at large scale (1M images), including challenging visual scenarios with artistic styles and fine-grained categories.

\paragraph{Superior data quality and scale:} Higher resolution, aesthetic quality, and category diversity with more instances per image and richer annotations aligned with natural image distributions.

\begin{figure}[H]
    \centering
    \includegraphics[width=1\linewidth]{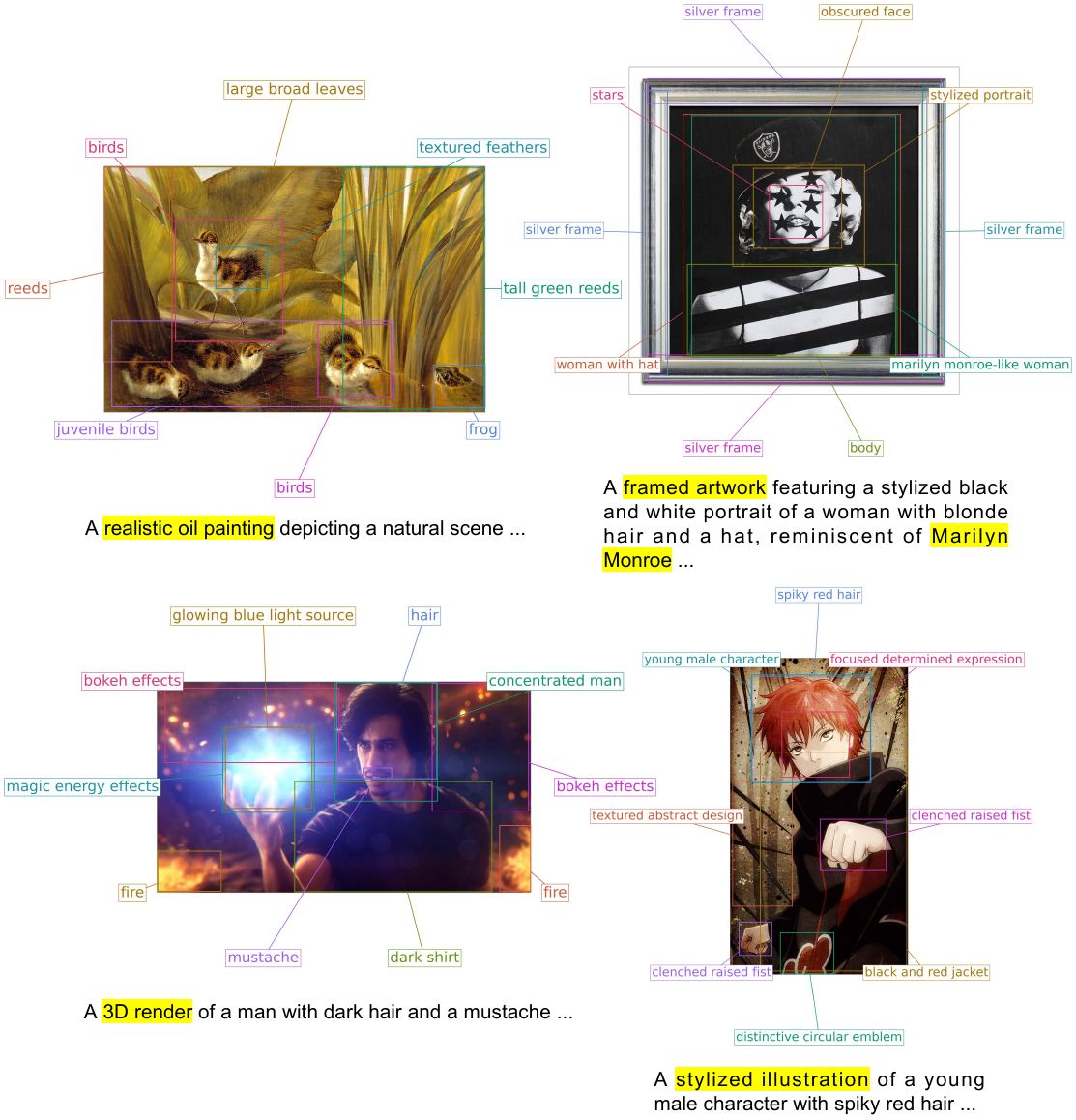}
    \caption{Our dataset encompasses stylized images from diverse internet sources, where 
    detailed VLM descriptions demonstrate particular advantages. The examples show successful 
    detection across various artistic styles, including oil paintings, black-and-white comics, 
    3D renderings, and anime. Each case captures primary objects while identifying focal points 
    within different aesthetic contexts. The VLM's ability to understand and summarize artistic 
    styles proves valuable for training generative models across diverse visual aesthetics, 
    maintaining annotation quality regardless of stylistic variations from photorealistic content.}
    \label{fig:stylized}
\end{figure}

\paragraph{Challenging category coverage:} Successfully encompasses complex visual scenarios across diverse domains, including artistic styles, fine-grained object categories, and contextually-dependent semantic descriptions. \figref{fig:stylized} provides some examples with different artistic styles.

\paragraph{Global-local coherence:} Annotations are linked to global image context through compositional elements, ensuring instance descriptions maintain contextual consistency for effective grounding. The text highlights in \figref{fig:resample_ovd} illustrate how the generated prompt references detected instances.

\newpage

\subsection{Representative Examples}

\newpage

\begin{figure}[H]
    \centering
    \includegraphics[width=0.8\linewidth]{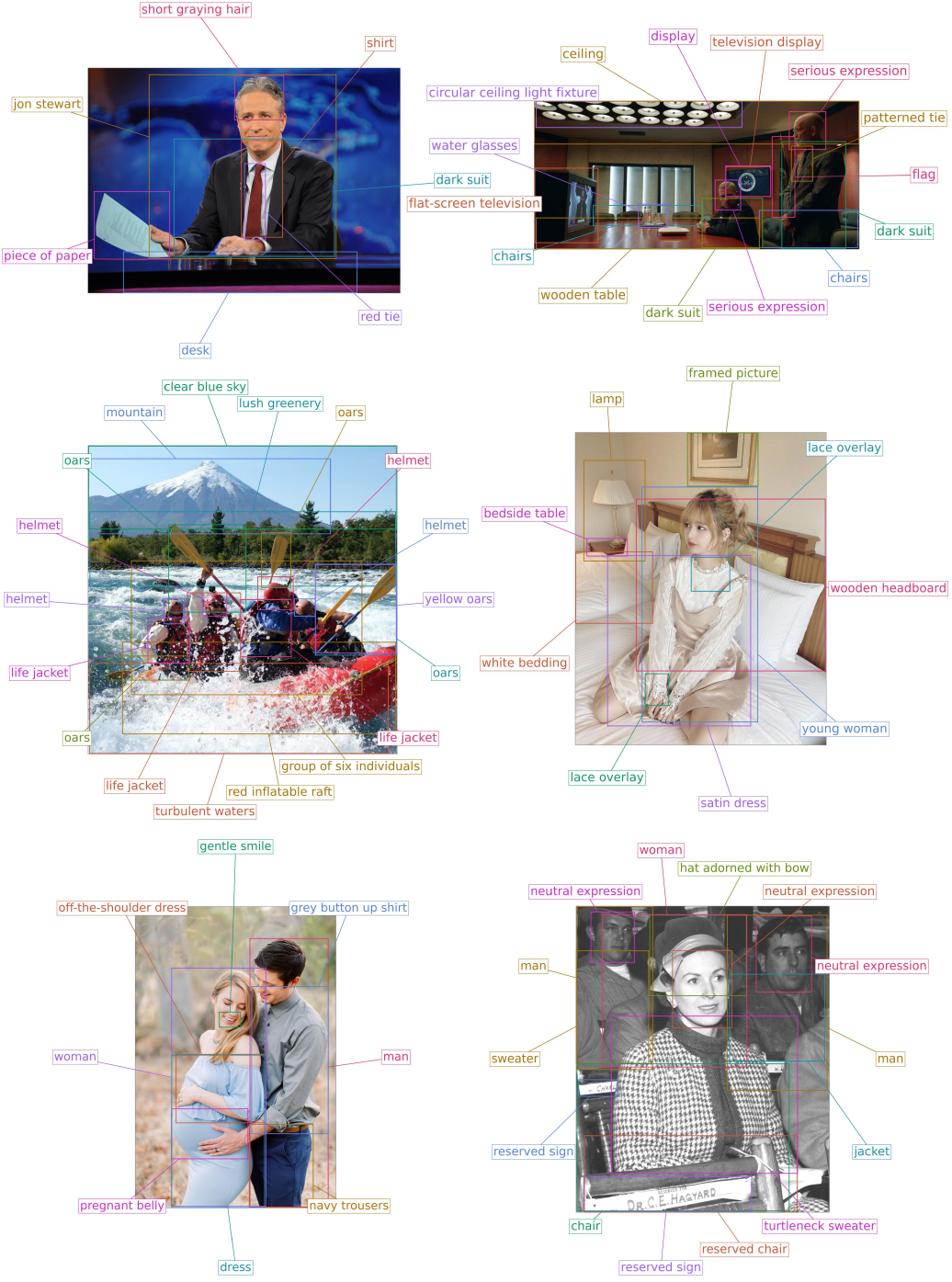}
    \caption{Human activities present significant challenges for traditional bounding box 
    detection approaches. Users consistently require detailed object descriptions for refined 
    generation outcomes, creating a gap between detection capabilities and generation needs. 
    Our dataset demonstrates progress in addressing this challenge through open-vocabulary 
    categories that facilitate detailed descriptions of human emotions, attire, and behavioral 
    states. The examples show how our approach captures nuanced human expressions and contextual 
    activities that standard detection vocabularies cannot adequately represent, enhancing the 
    descriptive richness for generative applications.}
    \label{fig:human_activity}
\end{figure}

\newpage

\newpage

\begin{figure}[H]
    \centering
    \includegraphics[width=1\linewidth]{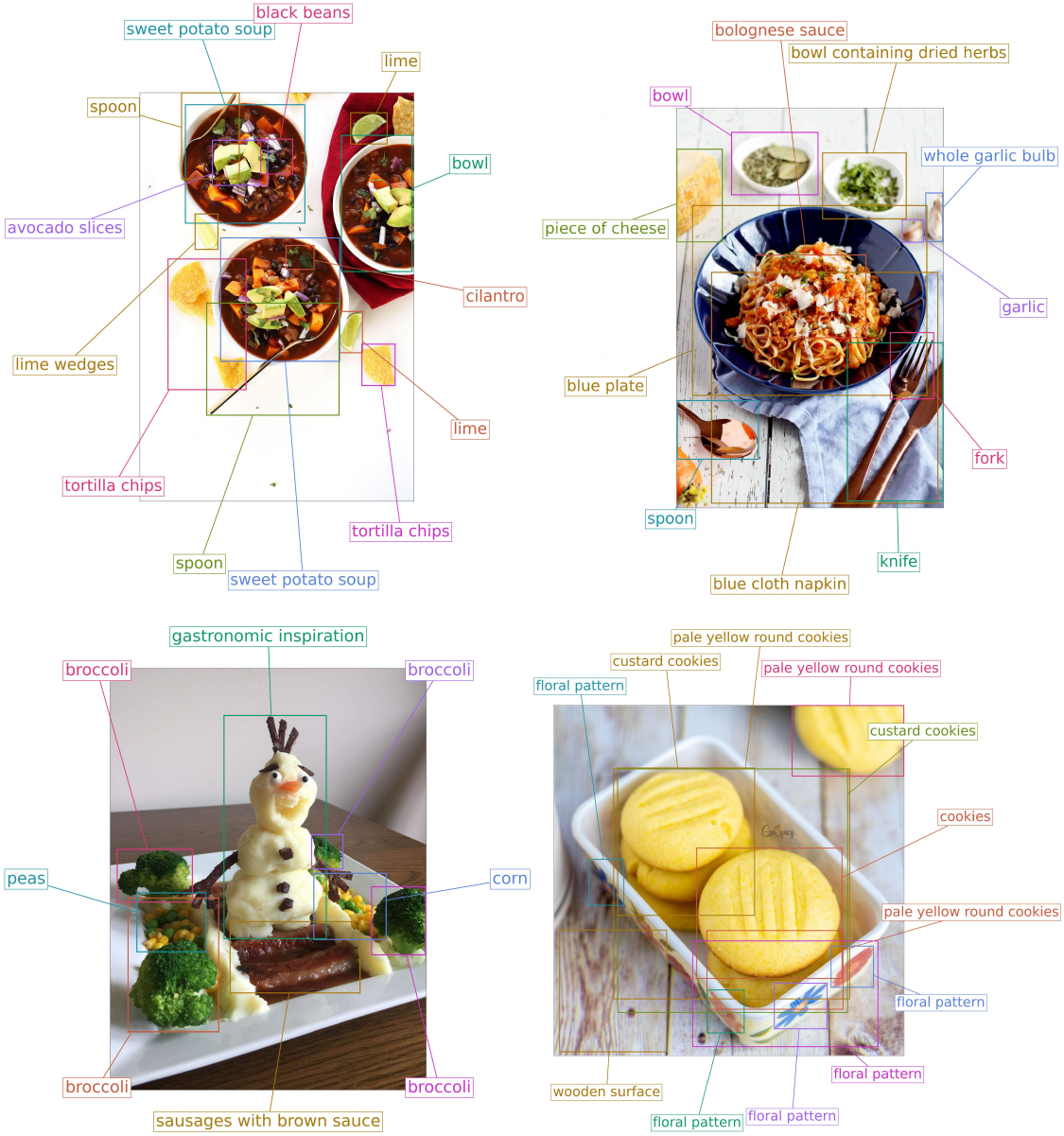}
    \caption{Food categories present particular complexity in object-centric detection and 
    constitute a significant proportion of social media images. Traditional detection methods 
    encounter challenges with food imagery, often producing outputs that are either excessively 
    dense or too sparse for effective generative model training. Our open-vocabulary framework 
    demonstrates benefits through specific categorization such as \emph{whole garlic bulb} in 
    the upper right and \emph{sausage with brown sauce} in the lower left. These examples 
    illustrate how detailed food categorization captures culinary specifics that generic 
    detection approaches typically miss, providing more precise semantic understanding for 
    food-related content.}
    \label{fig:foods}
\end{figure}

\newpage

\newpage
\section{\centering Aims of Resampling}\label{sec:resample}

\begin{figure}[H]
    \centering
    \includegraphics[width=0.9\linewidth]{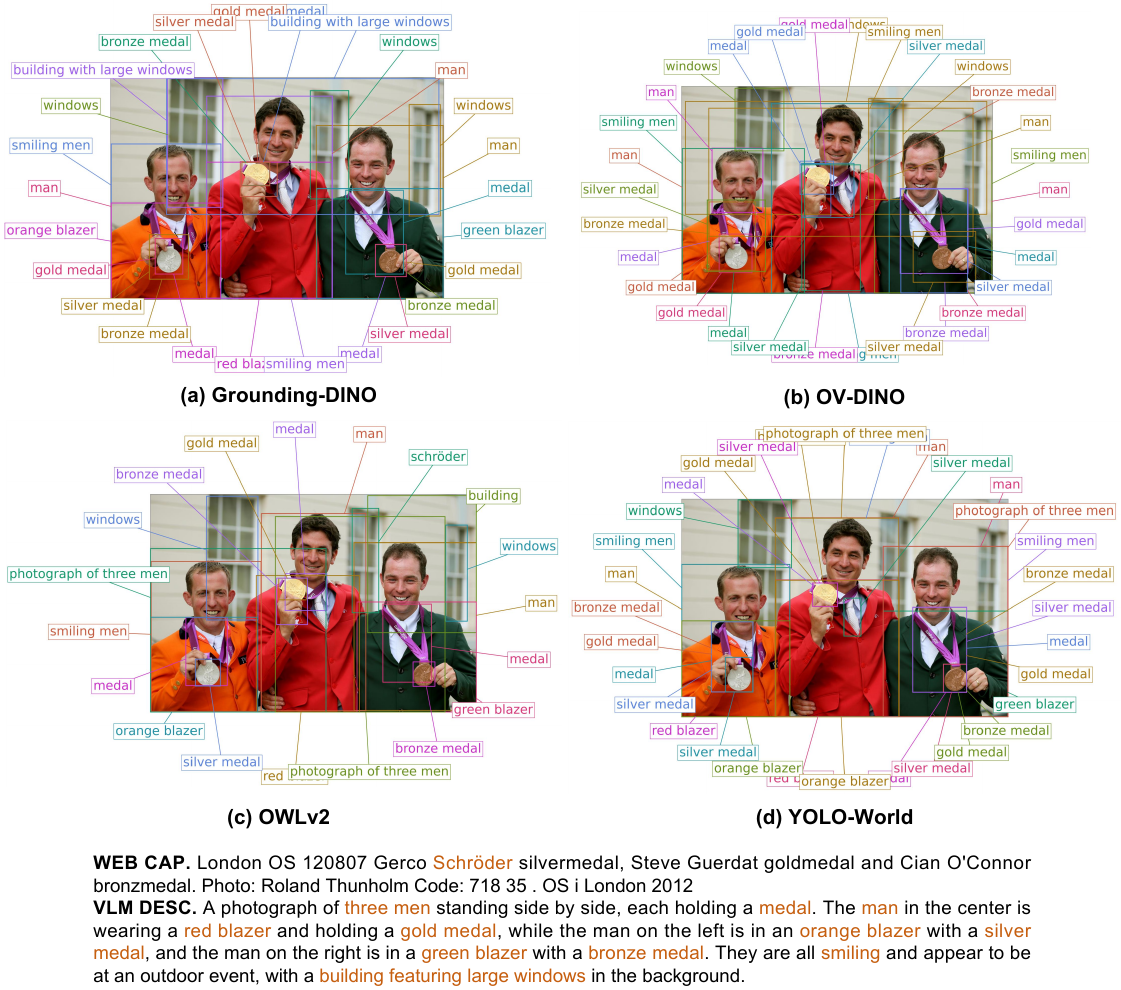}
    \caption{Effect of OVDs after the pre-detection VLM-LLM re-captioning.}
    \label{fig:resample_ovd}
\end{figure}

This section discusses the resampling stage mentioned in the paper's main text. Referring to \figref{fig:resample_ovd}, we first list the results of VLM description and LLM summarization for the categories to be detected; relevant words or phrases caught by OVDs~\cite{groundingdino, ovdino, owlv2, yoloworld} are highlighted in the text. Things evident are:
\begin{itemize}
\item Boxes from OVDs are pretty dense and exhibit significant overlap, which may introduce considerable potential bias. Furthermore, if each box is individually checked by the VLM, it would significantly increase the computational overhead.
\item Instances of errors are also apparent, such as inaccuracies regarding the materials of gold, silver, and bronze medals, as well as the specific color of blazers.
\item Notably, some detectors have incomplete category responses; for example, OV-DINO~\cite{ovdino} failed to detect any categories related to blazers in this case.
\item Additionally, there are isolated detection results, such as OWLv2~\cite{owlv2} responding to \emph{schröder}, a category derived from a web caption that falls under NER (Named Entity Recognition). Given the actual capabilities of OVDs, this category is not exceptionally reliable when three individuals are present.
\end{itemize}

\newpage
\begin{figure}[H]
    \centering
    \includegraphics[width=1\linewidth]{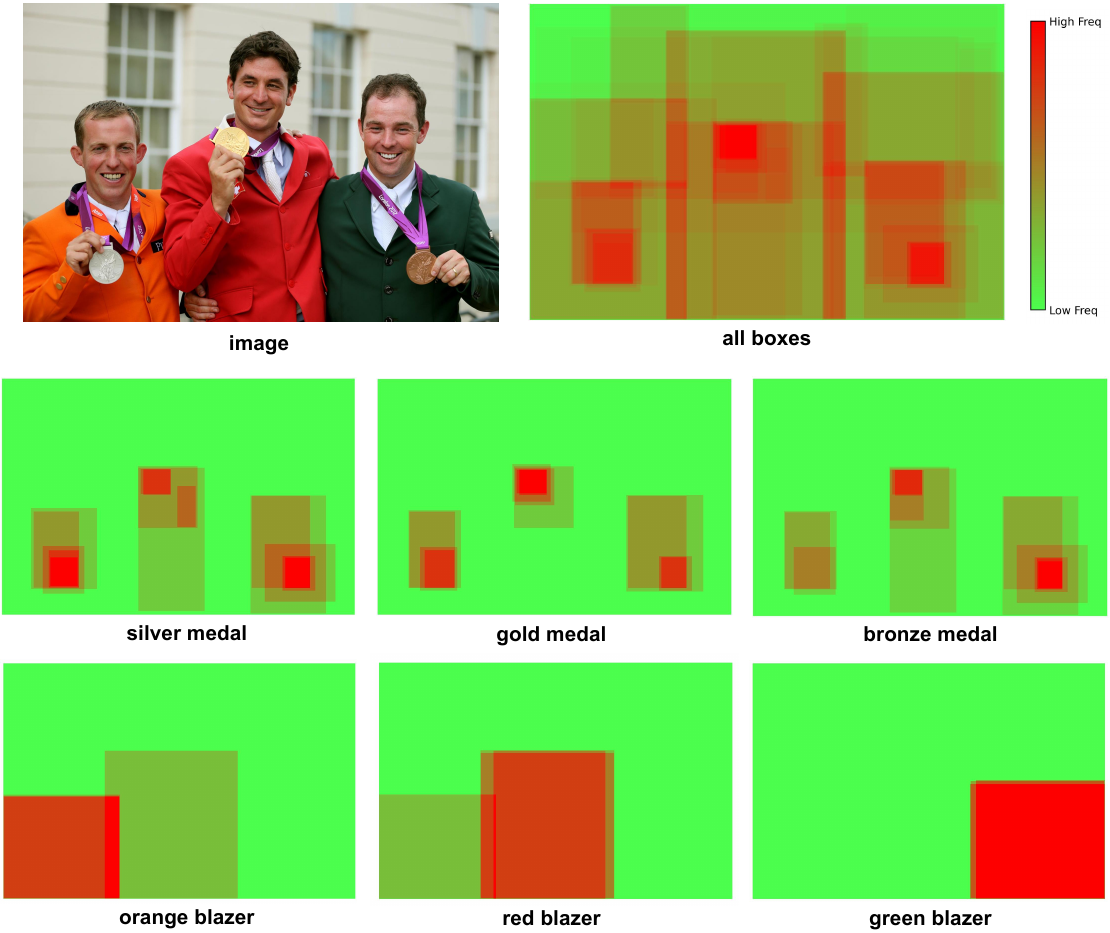}
    \caption{Heatmaps on categories of concern. }
    \label{fig:resample_heatmap}
\end{figure}

\begin{figure}[H]
    \centering
    \includegraphics[width=1\linewidth]{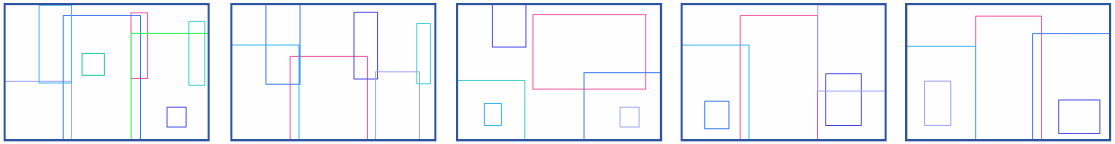}
    \caption{Box candidates in layers of resampling in relatively low overlap.}
    \label{fig:resample_layers}
\end{figure}

Here, we discuss how to integrate the detection results of OVDs from a fundamental perspective, specifically by examining the overlap relationships between bounding boxes. According to the frequency relationships revealed in \figref{fig:resample_layers}, the area containing the three medals exhibits the highest frequency. This observation also explains why we retain basic categories such as \emph{medal} and \emph{blazer} along with their open-vocabulary category combinations, as both reflect the visual observation capabilities of OVDs. 

We focus on the category combinations related to \emph{medal} and \emph{blazer}, as these are the most prone to errors. The heatmap indicates that we can determine priority based on overlap relationships by synthesizing the detection results from multiple OVDs. This results in the correct regions for these six groups of open-vocabulary categories achieving the highest frequency.

\newpage

\begin{figure}[H]
    \centering
    \includegraphics[width=1\linewidth]{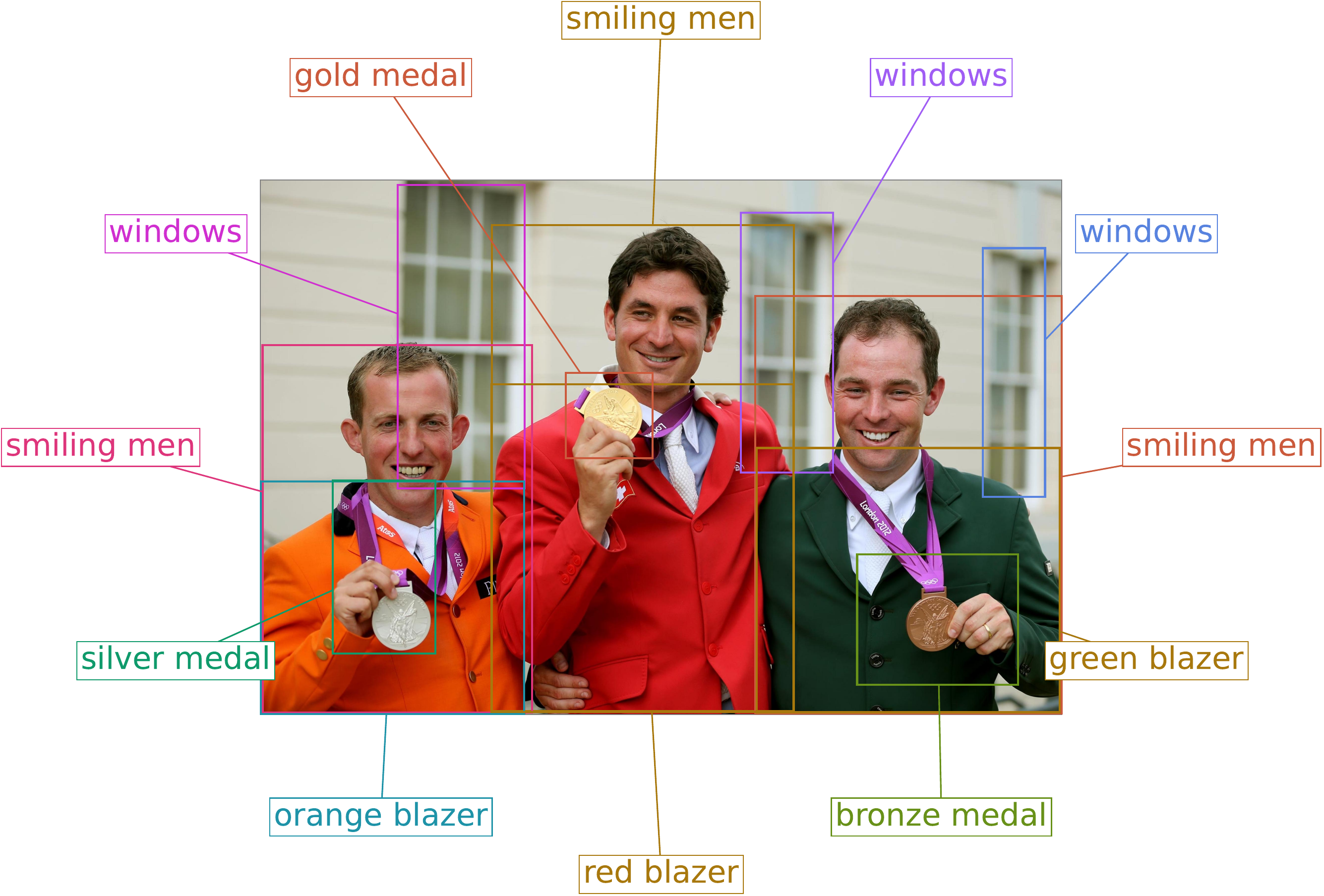}
    \caption{Effect of a limited number of resampled and cross-checked boxes.}
    \label{fig:resample_final}
\end{figure}

Regarding the regional overlap that is almost inevitable with OVDs, we identify a crucial factor: even in images with complex hierarchical relationships, the number of bounding boxes that can be effectively stacked for description remains limited. Therefore, when selecting candidate boxes for VLM inspection, we set our sampling target to five layers with lower overlap. The probability of sampling each instance is computed by penalizing several factors: previously sampled boxes, already sampled captions, distance from the image center, and small box sizes. Details of implementation could be found in the codes coming.

\figref{fig:resample_layers} illustrates the sampled layers, while \figref{fig:resample_final} encompasses the primary bounding boxes that successfully passed cross-checking by the VLM. This selection is sufficient to provide a comprehensive summary of the image. In this example, the original results from four OVDs yielded a total of 107 bounding boxes; however, only 30 boxes were sent for VLM cross-checking after resampling, with 21 ultimately passing. This approach significantly reduces potential information leakage while considerably conserving computational resources.

\newpage
\section{\centering Limitations in \OurDataset{}'s Annotations}\label{sec:limitation}

This section explores some seemingly questionable features and limitations within \OurDataset{} based on several typical examples.

\begin{figure}[H]
    \centering
    \includegraphics[width=1\linewidth]{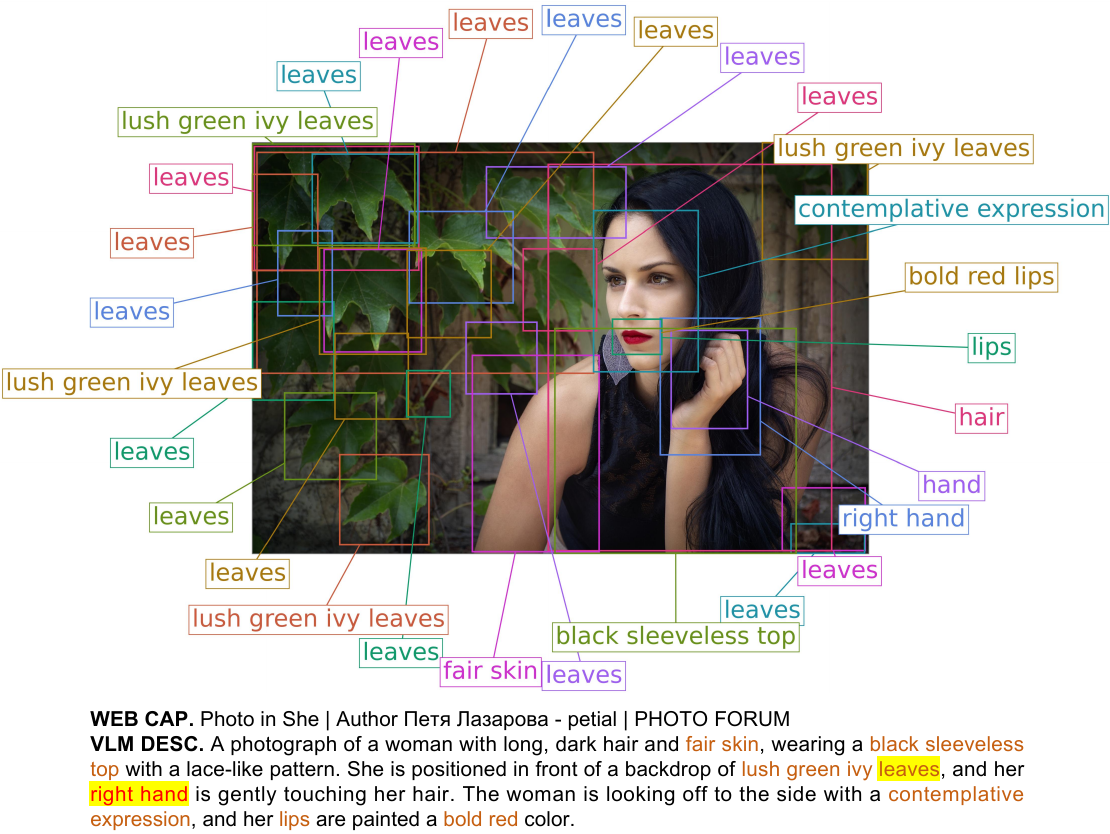}
    \caption{Effect of dense open-vocabulary detection based on VLM-LLM re-captioning.}
    \label{fig:limitation_woman}
\end{figure}

Based on \figref{fig:limitation_woman}, the following points can be analyzed:
\begin{itemize}
\item As demonstrated by \emph{leaves} or \emph{lush green ivy leaves}, we did not address singular and plural forms; instead, we generally maintained their original appearance in the VLM descriptions. This decision is based on several considerations: first, due to the inherent limitations of OVDs, if both ``leaves" and ``leaf" were treated as detection targets, the detector would struggle to differentiate between them. Second, our objective is to assist in text-to-image generation, where the text encoder, serving as a foundational module, can easily bridge the differences between singular and plural forms. Consequently, the training outcomes for text-to-image generation can broadly be shared across singular and plural forms based on the capabilities of the text encoder. As for inference, a more practical approach for users is to input multiple boxes in the singular form.
\item In the context of open-vocabulary, overlapping bounding boxes of different categories are almost inevitable. \secref{sec:resample} previously mentioned the potential misdetection issues related to medals of different materials; based on similar considerations, even when detecting "lush green ivy leaves," the original form "leaves" is not discarded.
\item The overlap among bounding boxes of the same category can be somewhat mitigated using techniques such as Non-Maximum Suppression (NMS), although we still retain some overlaps. This is particularly true for combinations of categories that exhibit weaker responses and present detection challenges, where it can be difficult to determine which bounding box is more appropriate or whether to include as many similar objects as possible within a single box.
\end{itemize}

\newpage
\begin{itemize}
\item \emph{right hand} is an error in the VLM description. We found that the VLM does not recognize a person's mirrored relationship within a photograph; thus, positional judgments such as ``right" are primarily based on the object's location within the image. Fortunately, such errors are quite rare and have minimal impact.
\end{itemize}

\begin{figure}[H]
    \centering
    \includegraphics[width=1\linewidth]{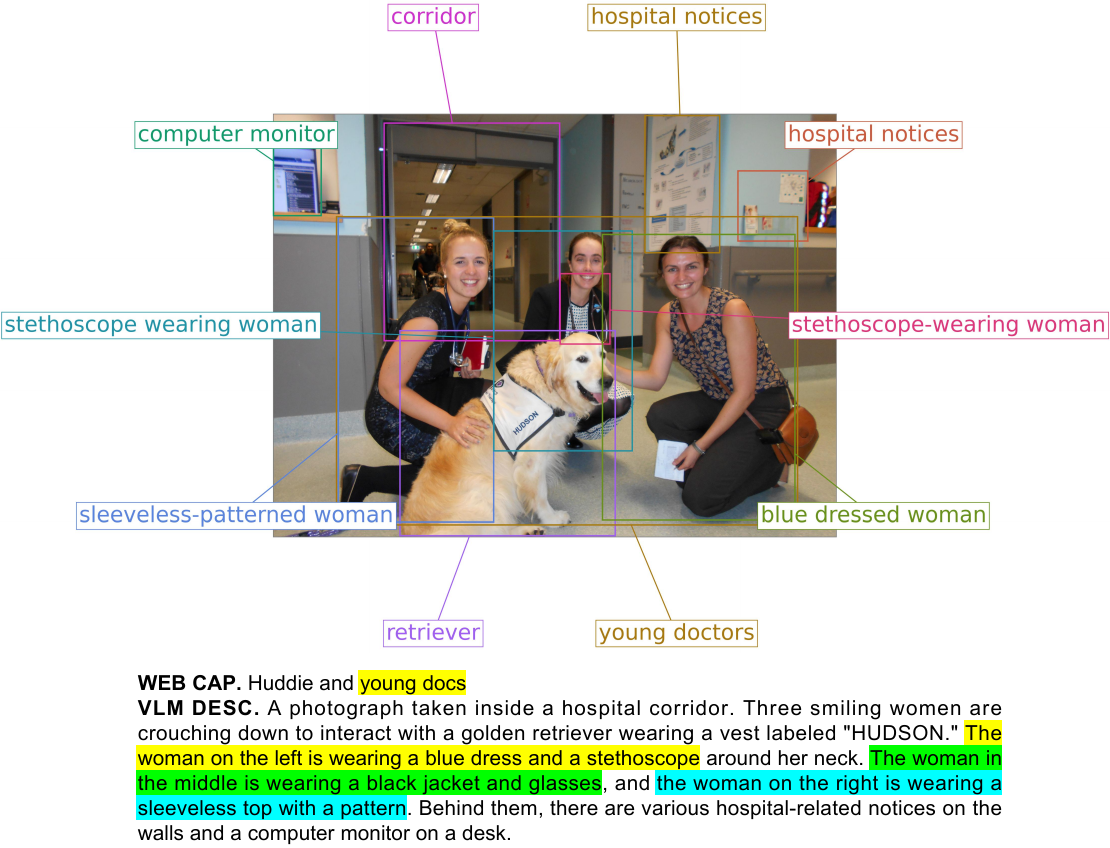}
    \caption{LLM summarized questionable open-vocabulary categories.}
    \label{fig:limitation_human_saying}
\end{figure}

\figref{fig:limitation_human_saying} illustrates instances where the LLM generates some unreasonable open-vocabulary categories, primarily related to human attire. Among these, \emph{stethoscope-wearing woman}, \emph{sleeveless-patterned woman}, and \emph{blue dressed woman} are not particularly rational categories.

Our experiments revealed that the LLM's tendency to synthesize categories across discontinuous text is generally beneficial; for example, \emph{bold red lips} in \figref{fig:limitation_woman} demonstrates this advantage. Although the category \emph{blue dressed woman} deviates somewhat from typical human conventions, it can still be utilized to some extent after processing through the text encoder. After weighing these factors, we accepted the LLM's inductive tendencies.

Another issue pertains to the misidentification of human identities. The example in the figure is derived from the web caption \emph{young doctor}. However, due to a lack of distinct visual elements, we cannot be certain whether the three women belong to this category (the woman wearing a stethoscope may be more closely aligned). Similar issues may arise with NER categories; for instance, in \figref{fig:resample_ovd}, OWLv2's detection results include \emph{Schröder}, but we cannot confirm which of the three awardees this individual represents.

\newpage
\begin{figure}[H]
    \centering
    \includegraphics[width=0.9\linewidth]{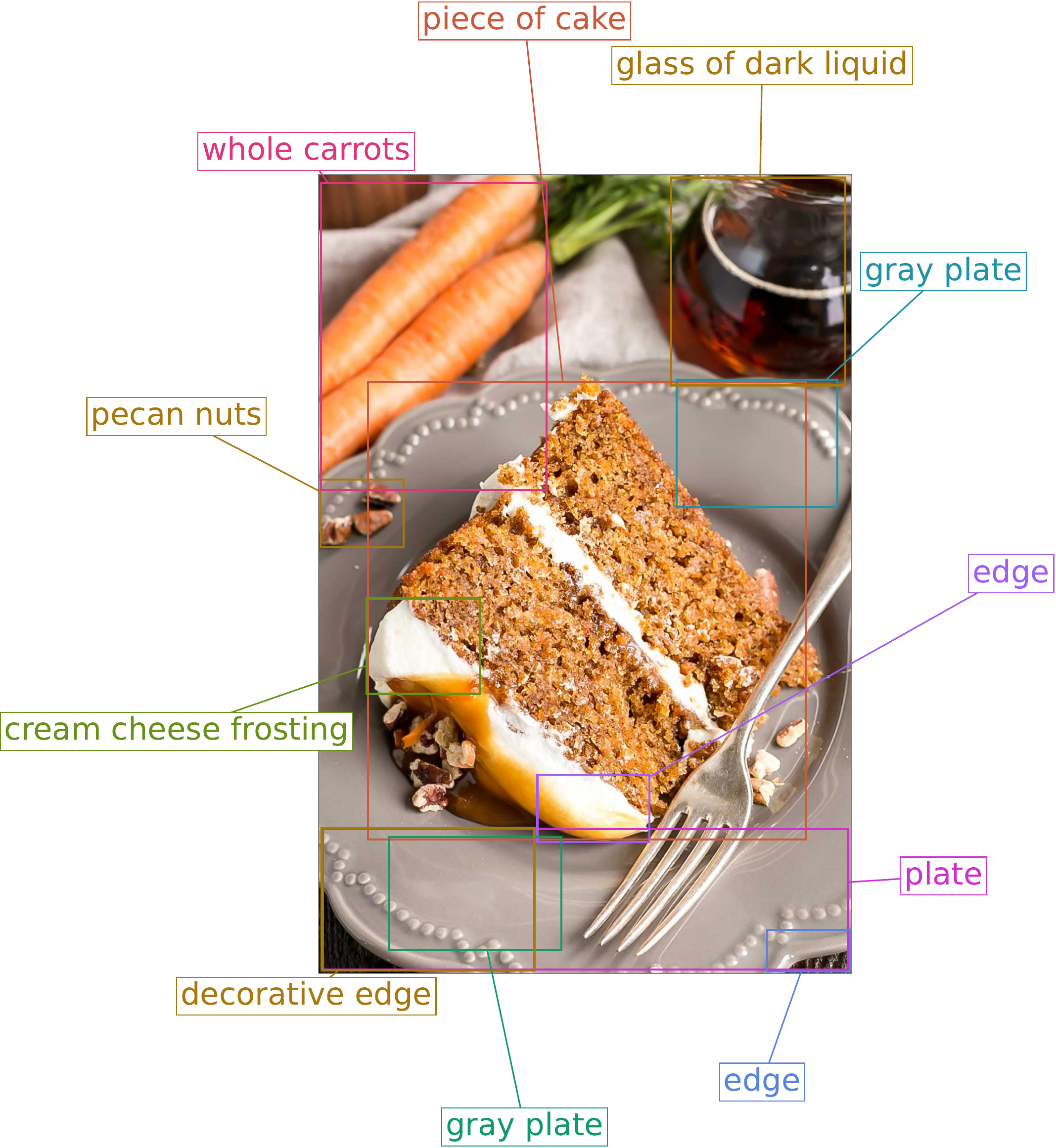}
    \caption{Discontinuous bounding box segments of \emph{edge} and \emph{gray plate}.}
    \label{fig:limitation_food_partial}
\end{figure}

\figref{fig:limitation_food_partial} illustrates the issue of discontinuous bounding boxes. Due to visual occlusion, the \emph{plate} and \emph{edge} in this example are not continuous, ultimately resulting in the detection results being fragmented into undesirable small segments. This problem arises from the tendencies inherited from OVDs.

Different OVDs may exhibit varying inclinations when faced with non-contiguous objects; they might either segment according to precise boundaries or attempt to summarize them into a cohesive whole that spans invisible portions, and sometimes they may exhibit both tendencies. The former can lead to overly fragmented bounding boxes, commonly seen in examples such as mountains, walls, water, flooring, etc. Conversely, the latter may result in overestimating size, as is often the case with items like a shirt worn underneath a coat.

Our approach is to leverage a resampling strategy that focuses on overlapping relationships to enhance candidate boxes' quality by utilizing knowledge from multiple OVDs. However, the results still fall short of expectations in a few instances.

\newpage
\section{\centering Failure Cases in Generation}\label{sec:failure}

\begin{figure}[H]
    \centering
    \includegraphics[width=0.75\linewidth]{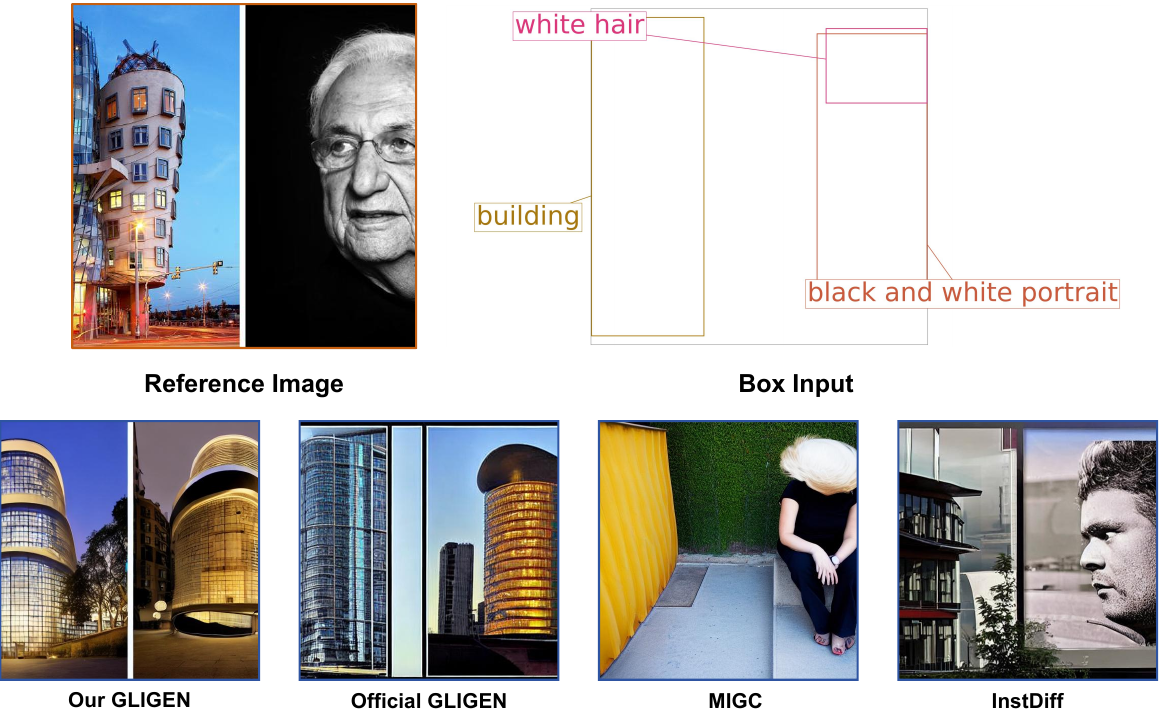}
    \caption{In the image above, the left side of the bounding box input represents a building, while the right side depicts a person, resembling a collage with two independent sections. Due to being trained on global VLM descriptions, our model places a strong emphasis on coherence, making it less likely to generate results that appear disjointed and lack natural transitions.
Comparative results indicate that official GLIGEN is similar to our model, whereas MIGC and InstanceDiffusion successfully generated mutually independent objects. This observation reflects the latter two models' focus on independent instance grounding.}
    \label{fig:failure_dual}
\end{figure}

\begin{figure}[H]
    \centering
    \includegraphics[width=0.75\linewidth]{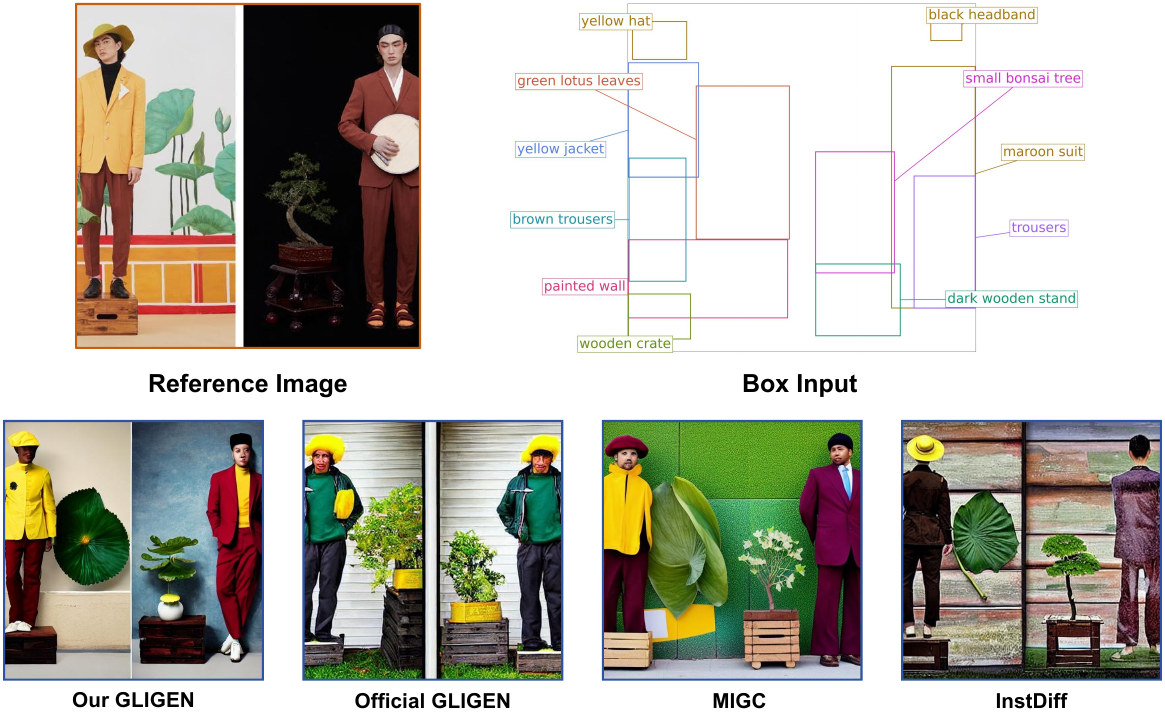}
    \caption{The category \emph{small bonsai tree} is a low-frequency open-vocabulary category within our dataset, which may hinder the potentials of corresponding generative capabilities. In this example, the official GLIGEN's generation results for ``bonsai" may be closer to expectations. However, the attribute binding for our other composite categories is noticeably superior, particularly in terms of color.}
    \label{fig:failure_bonsail}
\end{figure}

\newpage
\begin{figure}[H]
    \centering
    \includegraphics[width=0.9\linewidth]{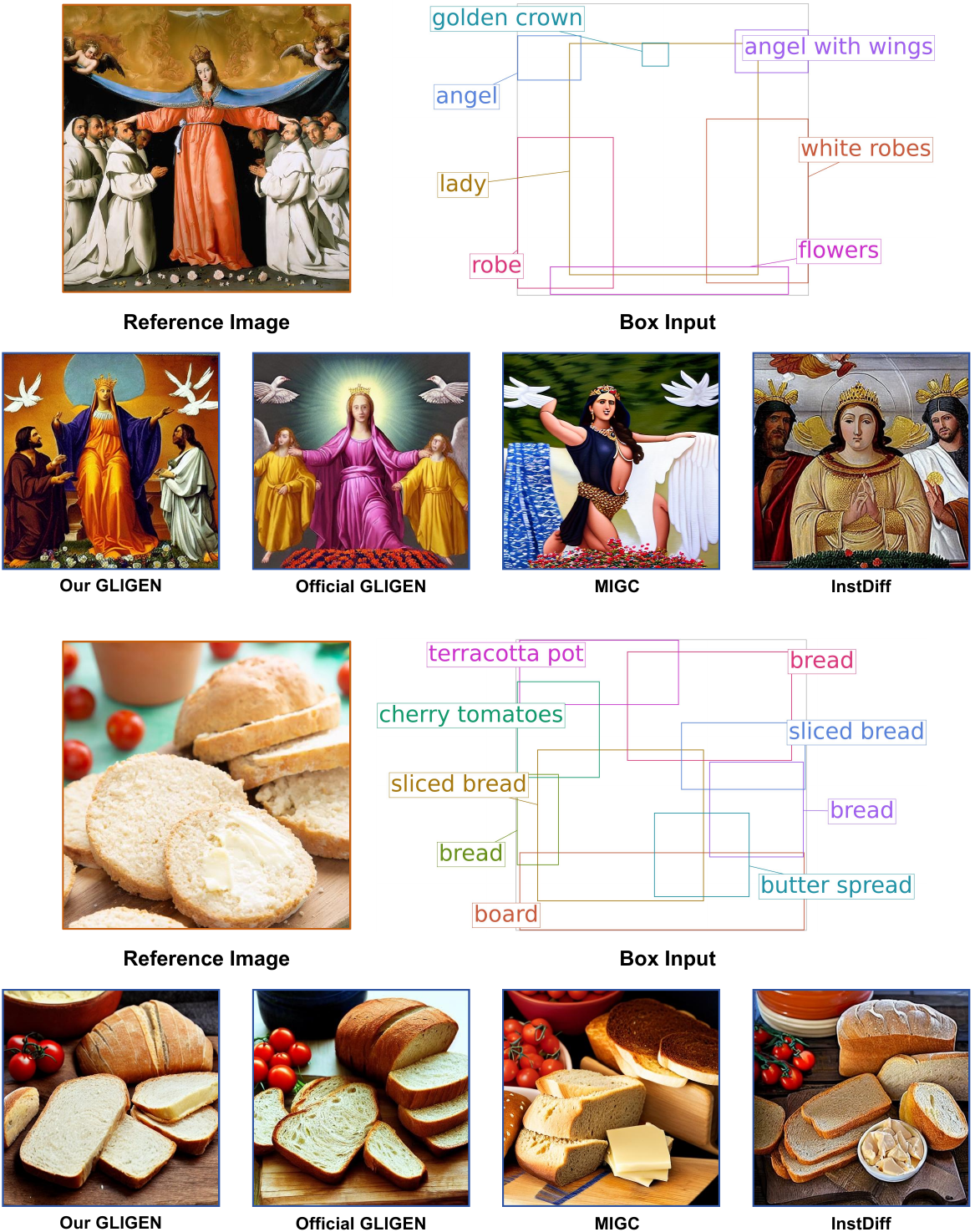}
    \caption{In this set of examples, the \emph{angel} in the upper group and the \emph{butter spread} in the lower group were not generated correctly. This is likely attributable to the relatively low-frequency visual associations present in the training dataset. For instance, all models in the upper group generated the bounding box labeled \emph{angel} as a bird, which may reflect a common trend observed in the network images from the training set. In the lower group, both MIGC and InstanceDiffusion produced independent instances of \emph{butter} that were unrelated to \emph{spread}, implying a trend focus on independent instance grounding, similar to that seen in \figref{fig:failure_dual}.}
    \label{fig:failure_ov}
\end{figure}

\newpage
\begin{figure}[H]
    \centering
    \includegraphics[width=0.9\linewidth]{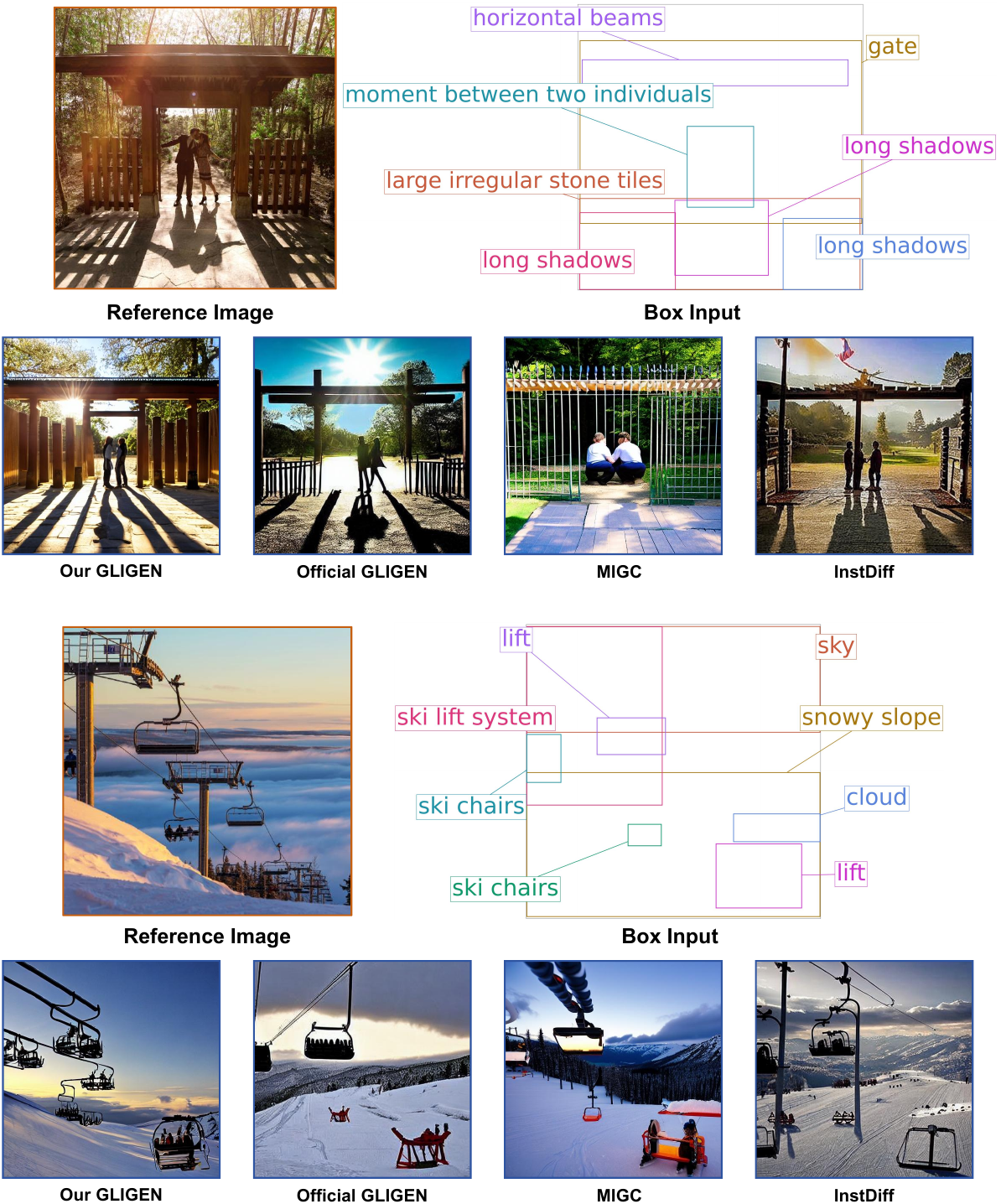}
    \caption{This set of examples illustrates several visual elements that extend beyond the spatial representation capabilities of bounding boxes. In the upper group, \emph{shadows} are clearly better represented through segmentation masks, while merely providing bounding boxes significantly challenges the logical capabilities of the base model itself. In the lower group, the examples related to the \emph{ski lift system} resulted in a complete failure of generation, indicating that isolated bounding boxes alone are insufficient to guide the model in capturing the suspended state of the cable cars along continuous tracks.}
    \label{fig:failure_segment}
\end{figure}

\newpage
\section{\centering More Information about User Study}\label{sec:userstudy}

\begin{figure}[h]
    \centering
    \includegraphics[width=1\linewidth]{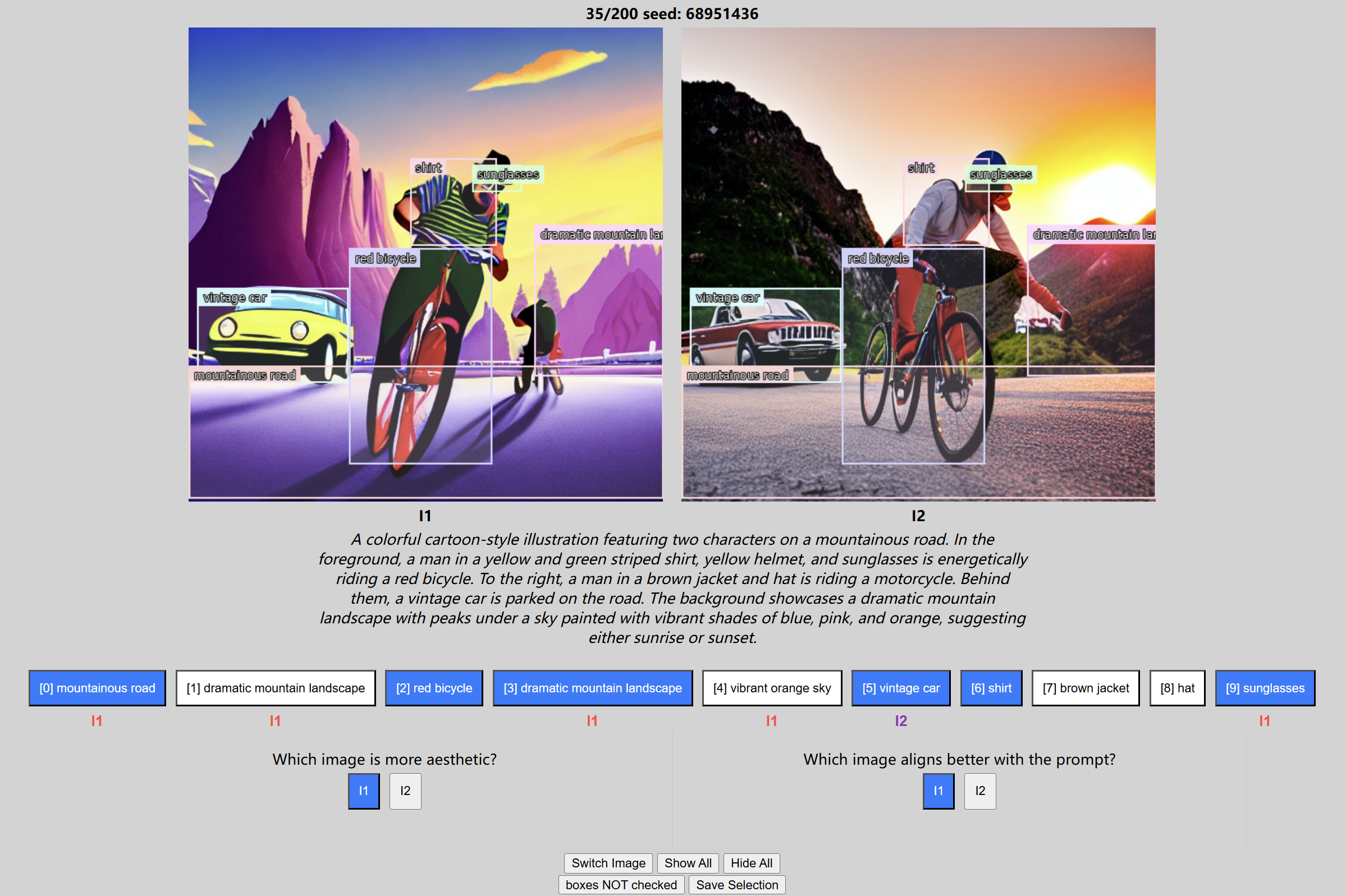}
    \caption{A screenshot of our user study interface.}
    \label{fig:user_study}
\end{figure}

\figref{fig:user_study} presents a screenshot of our evaluation interface, which shows a randomized pair of generated images alongside the input prompt and labeled bounding boxes. Participants were asked to choose a winner among the generated pairs based on: (1) aesthetic quality and (2) alignment with the provided prompt, without knowing the source generator. Optionally, participants could select a better instance alignment for each individual box-label pair. Users could toggle the bounding box overlays to examine occluded content. Most operations could be performed using either buttons or hotkeys.

Our initial assessment revealed that evaluating box-label alignment precision and complex prompt adherence (derived from VLM descriptions) was cognitively demanding, due to the large amount of text and boxes that have to be processed for each image. Considering the effort required and to prevent potential VLM misuse during evaluation, we prioritized evaluator expertise and trustworthiness. Due to the voluntary nature of this study, we recruited a group of colleagues experienced in image generation whom we could trust to complete the evaluations without resorting to LLM/VLM assistance.

Prompt alignment and aesthetic quality were assessed at the image level, while instance alignment was evaluated at the object level—comparing each box-label pair with its corresponding region in the generated images. This approach allowed us to measure both overall image coherence and the precision of individual object grounding within the synthesized outputs.

\newpage
\twocolumn
\section*{Acknowledgments}
This work is supported by NSF China (No. 62421003) and the XPLORER PRIZE.

{
    \small
    \bibliographystyle{ieeenat_fullname}
    \bibliography{main}
}

\end{document}